\pdfoutput=1
\documentclass{article}
\usepackage{multirow}
\usepackage{graphicx}
\usepackage{amsmath}
\usepackage{colortbl}
\usepackage{bbding}
 \usepackage[preprint]{neurips_2026}

\usepackage[utf8]{inputenc} 
\usepackage[T1]{fontenc}    
\usepackage{hyperref}       
\usepackage{url}            
\usepackage{booktabs}       
\usepackage{amsfonts}       
\usepackage{nicefrac}       
\usepackage{microtype}      
\usepackage{xcolor}         

\title{PathoSage: Towards Multi-Source Evidence Adjudication in Pathology via Experience-Aware Agentic Workflow}

\author{%
Chengyang Zhang$^{1,2}$\thanks{Equal Contribution} \quad Wenchuan Zhang$^{2*}$ \quad Bo Li$^3$ \quad Mengran Li$^4$  \quad Bob Zhang$^3$\\
\textbf{Yuhao Yi}$^{1,2}$\thanks{Corresponding Author} \quad \textbf{Hong Bu}$^{2,\dag}$ \quad \textbf{Jiancheng Lv}$^{1,\dag}$\\
$^1$College of Computer Science, Sichuan University \\
$^2$Department of Pathology and Institute of Clinical Pathology, West China Hospital, Sichuan University \\
$^3$Department of Computer and Information Science, University of Macau\\
$^4$School of Intelligent Systems Engineering, Sun Yat-sen University\\
\texttt{yuhaoyi@scu.edu.cn}\\
}

\begin{document}

\maketitle

\begin{abstract}
  Recent advances in Multimodal Large Language Models (MLLMs) and agent workflows have shown strong promise for computational pathology, yet reliable patch-level reasoning remains challenging. End-to-end pathology MLLMs often hallucinate morphological features, while recent agentic systems usually merge tool outputs and retrieved knowledge into a shared context, making decisions vulnerable to conflicting evidence and context contamination. We propose PathoSage, a three-stage framework that explicitly separates knowledge retrieval, evidence collection, and evidence adjudication for patch-level pathology multimodal reasoning. Its core component, Structured Evidence Deliberation, independently evaluates heterogeneous evidence from tools, performs conflict analysis, and generates the final judgment in a fresh context to reduce anchoring bias. We further introduce a training-free Beta-Bernoulli experience system with continuous credit assignment to model long-term tool reliability and construct similarity-weighted priors for future tool use. Experiments show that PathoSage effectively mitigates VQA hallucinations and classifier disagreement, outperforming strong pathology MLLM and agentic baselines. Our results highlight explicit evidence adjudication and reliability-aware tool modeling as key ingredients for robust pathology agents.
\end{abstract}

\section{Introduction}
In recent years, Multimodal Large Language Models (MLLMs) in computational pathology have rapidly advanced from early image-text representation learning to complex multi-step reasoning~\cite{plip,quilt1m,slideseek,vlsa}. Consequently, pathology AI is evolving from a monolithic paradigm toward agentic systems that actively invoke external tools, retrieve domain knowledge, and organize analysis workflows. Moving beyond direct answer generation, these pathology agents increasingly emulate expert behavior by utilizing structured mechanisms to acquire and organize evidence~\cite{pathchat,wsi-llava,patho-agenticrag,cpathagent,pathology-cot}. Recent studies have advanced multimodal reasoning across both fine-grained morphological recognition and whole-slide image (WSI) cross-region analysis~\cite{titan,chief}. Tool augmentation has also emerged; for example, PathAsst integrated a specialized backbone with visual sub-models and literature retrieval~\cite{pathasst}. Building on this, recent studies highlight that reliable reasoning requires structured workflows for observation selection, tool invocation, and progressive evidence accumulation, rather than relying solely on stronger visual representations~\cite{patho-agenticrag,cpathagent,pathology-cot}. Taken together, these developments suggest that pathology AI is gradually shifting from isolated multimodal understanding toward more structured systems that must organize, compare, and utilize evidence across multiple sources.

Despite these advances, existing methods typically merge tool outputs, retrieved information, and model reasoning into a single shared context~\cite{Rajendran2025FoundationMI, peng2025aligning}. This design is fragile when sources provide \emph{heterogeneous or conflicting evidence}, such as disagreeing classifiers, hallucinated VQA findings, or misaligned retrieved knowledge~\cite{chen2026landscape}. Without explicit evidence adjudication, early biases and context contamination accumulate, reducing reliability and interpretability. The core challenge, therefore, is not merely how to add more tools or more knowledge, but how to separate, assess, and reconcile heterogeneous evidence before producing a final answer. This limitation mirrors broader challenges recognized in literature on reasoning agents, tool use, and retrieval-augmented generation~\cite{react,toolformer}.
This issue is particularly critical for \emph{patch-level pathology reasoning}. As compact and interpretable units of morphological evidence~\cite{zhang2025attention,shui2026nunext}, local patches serve as a natural foundation for clinical judgments, educational assistance, and interactive analysis~\cite{conch,musk,homie,patho-agenticrag,octomed,pulsemind,cx-mind,wu2025bridging,jeddi2026does,anatomy-r1}. While the patch-level setting provides an ideal testbed for studying tool-model interactions, there remains no unified framework for structurally collecting, reconciling, and modeling heterogeneous evidence. It offers a relatively controlled setting in which the central difficulty is not large-scale navigation itself, but how multi-source evidence should be organized, compared, and adjudicated. The fundamental question of \emph{how multi-source evidence should be adjudicated} remains underexplored~\cite{rag,reflexion,zhang2026multimodal}.

\begin{figure*}[t]
  \includegraphics[width=\textwidth]{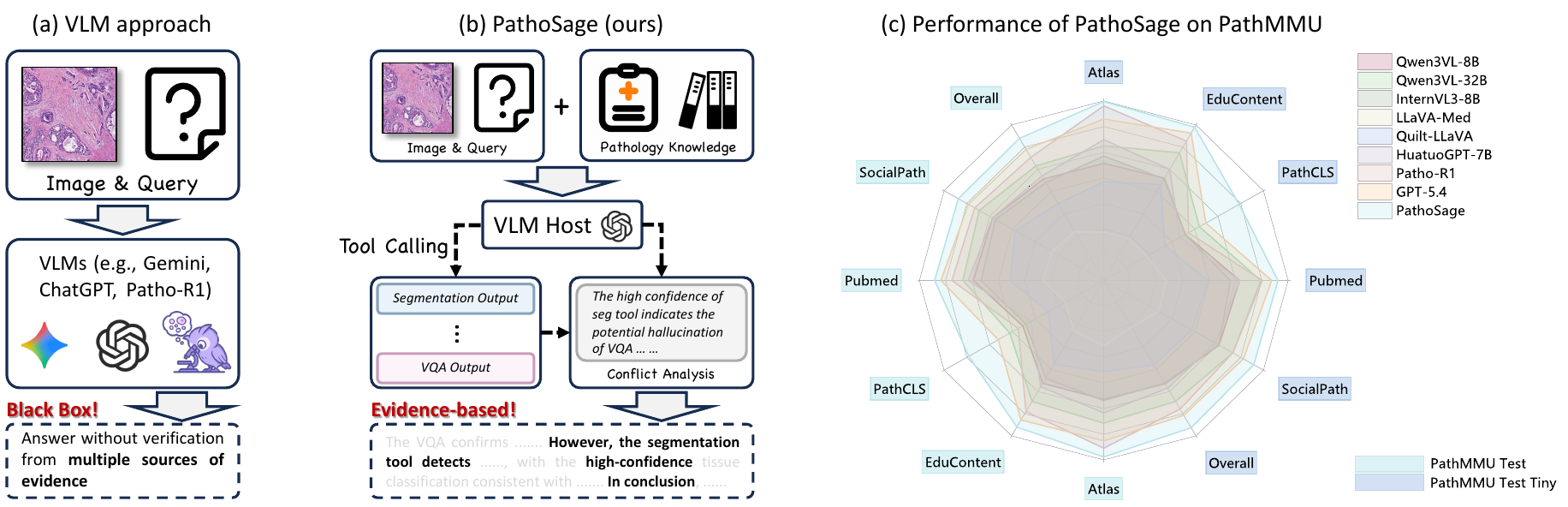}
  \caption{Comparison of the (a) "black box" VLM approach and (b) our proposed PathoSage for evidence-based pathology analysis. (c) is the performance of PathoSage on the PathMMU test set.}
  \label{fig1}
\end{figure*}

To address this, we propose PathoSage, a three-stage framework for patch-level multimodal reasoning that explicitly decomposes the process. First, in the knowledge retrieval stage, the system retrieves and assesses task-relevant external knowledge based on the patch and query~\cite{patho-agenticrag,pathasst}. Next, the evidence collection stage invokes pathology-specific tools to gather local visual evidence, deferring the final answer generation~\cite{react}. Finally, the Structured Evidence Deliberation stage independently evaluates tool outputs, performs conflict analysis, and generates a final judgment in a fresh context to minimize historical contamination. Thus, PathoSage shifts the paradigm from merely \emph{using} tools to explicitly \emph{adjudicating} their evidence.
Furthermore, we introduce a Beta-Bernoulli experience system to dynamically model tool reliability across similar patches~\cite{agrawal2012analysis}. Rather than assuming static trustworthiness, PathoSage continuously updates posterior estimates based on tool performance and task relevance, bridging single-instance reasoning with long-term adaptation for more targeted future tool use~\cite{toolmem,xskill}. Ultimately, by formalizing evidence organization and adjudication, this work establishes a robust foundation for both practical patch-level applications and larger-scale pathology agent systems (Fig.~\ref{fig1}).
\paragraph{Our main contributions are as follows.}
\begin{enumerate}
\item We propose \textbf{PathoSage}, a three-stage agent framework for patch-level pathology multimodal reasoning that explicitly decouples knowledge retrieval, evidence collection, evidence adjudication, and final answer generation.
\item We introduce \textbf{Structured Evidence Deliberation (SED)} and a \textbf{Beta-Bernoulli experience system} for heterogeneous evidence assessment, inter-tool conflict analysis, weighted reasoning, and long-term reliability-aware tool utilization.
\item We build a tool-augmented system for patch-level reasoning, validating it across multiple benchmarks to demonstrate the value of explicit evidence adjudication and experience-based reliability modeling.
\end{enumerate}

\section{Related Works}
\subsection{Pathology Multimodal Large Language Models}

In recent years, pathology multimodal large language models have advanced rapidly, with the research focus expanding from early image-text representation learning to pathology question answering, description generation, interpretability, and more complex multi-step reasoning. A natural trend in this evolution is that some studies primarily center on local pathology images, emphasizing fine-grained morphological recognition, local semantic understanding, and patch-level question answering, while others further extend to whole-slide images, modeling cross-region context, multi-scale tissue organization, and slide-level semantic generation. Representative works along these directions include PathAsst~\cite{pathasst}, Quilt-LLaVA~\cite{quilt-llava}, PathChat~\cite{pathchat}, PA-LLaVA~\cite{pa-llava}, PathGen-LLaVA~\cite{pathgen16m}, Patho-R1~\cite{patho-r1}, SmartPath-R1~\cite{smartpath-r1}, and TeamPath~\cite{teampath}, as well as WSICaption~\cite{wsicaption}, WSI-VQA~\cite{wsi-vqa}, HistGen~\cite{histgen}, SlideChat~\cite{slidechat}, WSI-LLaVA~\cite{wsi-llava}, PathAlign~\cite{pathalign}, ALPaCA~\cite{alpaca}, TITAN~\cite{titan}, PathReasoner-R1~\cite{pathreasoner-r1}, CPath-Omni~\cite{cpath-omni}, PolyPath~\cite{polypath}, HistoGPT~\cite{histogpt}, PRISM2~\cite{prism2}, Hepato-LLaVA~\cite{hepato-llava} and PathFound~\cite{pathfound}. Overall, existing pathology MLLMs have demonstrated that pathology understanding cannot rely on a single scale or modality alone, but instead requires connecting local morphological cues with higher-level histopathological semantics. 

\subsection{Tool-Augmented Reasoning and Pathology Agents}
As pathology multimodal systems continue to evolve, an increasing number of studies have shifted the focus from simply enabling models to answer questions toward enabling systems to actively organize the reasoning process. This trend is typically reflected in the introduction of agentic capabilities such as tool invocation, knowledge retrieval, region navigation, multi-step observation, and decision trajectory modeling. Unlike traditional pathology MLLMs, which mainly emphasize end-to-end generation, pathology agents place greater emphasis on whether the system can more closely mimic the workflow of pathologists by actively selecting regions of interest, invoking auxiliary modules, and progressively accumulating evidence before reaching a conclusion. Recent pathology agent research has already expanded to multiple directions, including knowledge-augmented reasoning, whole-slide navigation, clinical decision support, prognostic analysis, and biomarker discovery. Representative systems include Patho-AgenticRAG~\cite{patho-agenticrag}, SlideSeek~\cite{slideseek}, Pathology-CoT~\cite{pathology-cot}, PathFinder~\cite{pathfinder}, PathAgent~\cite{pathagent}, CPathAgent~\cite{cpathagent}, SurvAgent~\cite{survagent}, WSI-agent~\cite{wsi-agent}, TissueLab~\cite{tissuelab}, MMNavAgent~\cite{mmnavagent}, as well as related agent frameworks for oncology decision-making and biomarker discovery~\cite{ferber2025development, sage}. Overall, these studies suggest that pathology AI is evolving toward active systems that integrate tool use, knowledge access, and evidence accumulation. This shift also highlights a deeper challenge: how heterogeneous evidence should be organized and used for reliable reasoning.

\subsection{Multi-Source Evidence Integration, Conflict Handling, and Reliability Modeling}

Although tool augmentation, retrieval-augmented generation, and agentic reasoning have substantially expanded the capability boundary of multimodal systems, most existing methods still combine tool outputs, retrieved knowledge, and model reasoning within a shared interaction trajectory, leaving the final decision to be made over a single accumulated context~\cite{react,toolformer,rag,pathasst,cpathagent,pathology-cot}. While this design is effective for improving overall capability, it raises an important and still underexplored challenge: when different tools provide heterogeneous, partially relevant, or even conflicting evidence, how should a system explicitly separate \emph{evidence collection} from \emph{evidence adjudication}? This issue is particularly critical in pathology, where classifiers may disagree, VQA modules may hallucinate morphological findings, and retrieved knowledge may only partially align with the image under analysis.
Recent studies have begun to move beyond one-shot tool use toward memory- and experience-aware agents~\cite{memos,memverse}. ToolMem shows that agents can improve tool selection by summarizing the strengths and weaknesses of tools from prior interactions and retrieving such capability memory at inference time~\cite{toolmem}. XSkill further highlights the importance of continual learning from both \emph{experiences} and \emph{skills}, using visually grounded summarization, cross-rollout critique, and retrieval-based adaptation to improve multimodal agents without parameter updates~\cite{xskill}. However, these methods mainly focus on improving tool selection and continual adaptation, rather than explicitly modeling how heterogeneous tool outputs should be independently assessed, reconciled under conflict, and translated into reliability-aware final decisions in pathology reasoning.

\begin{figure*}[t!]
  \centering
   \includegraphics[width=1.\textwidth]{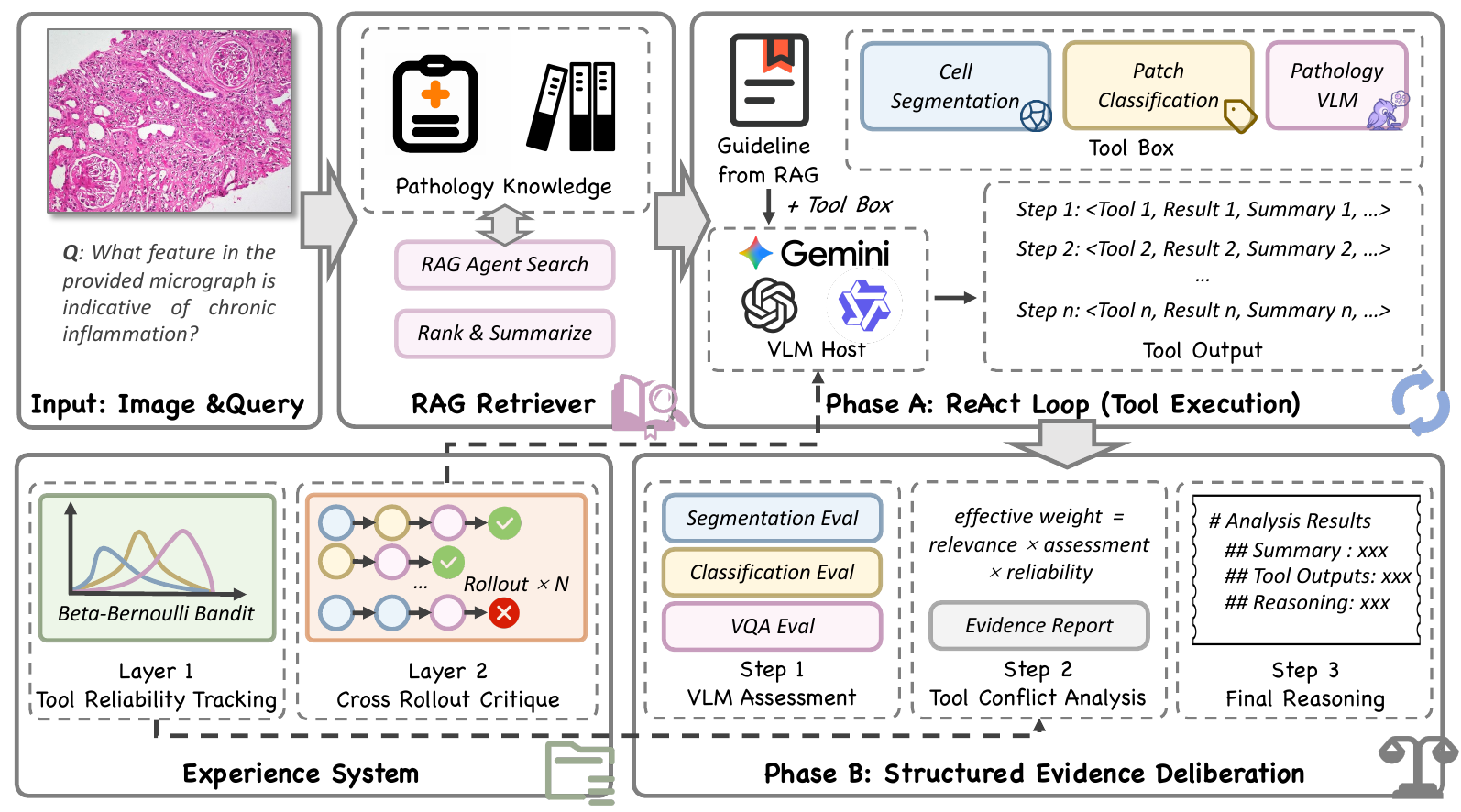}
   \caption{Overview of the PathoSage framework, which performs pathology multimodal reasoning through a three-stage agentic system with an optional experience system.}
   \label{fig2}
\end{figure*}

\section{Methods}
\subsection{Overview}
Given a pathology image $I$ and a clinical query $Q$, PathoSage uses a set of specialist tools $\mathcal{T}=\{t_1,\dots,t_m\}$ to produce an answer $\mathcal{A}$. The central problem is the adjudication of multi-source evidence: individual tools may return complementary yet contradictory conclusions, and naively presenting all outputs to a single VLM call induces anchoring bias \cite{echterhoff2024cognitive}, which degrades fusion quality. To address this, PathoSage decouples the analysis into three phases with strict information isolation (Fig.~\ref{fig2}): a RAG-based knowledge retrieval stage provides domain-grounded tool planning; a ReAct loop collects tool evidence; and a Structured Evidence Deliberation (SED) procedure independently assesses, algorithmically weighs, and synthesizes the collected evidence. A Bayesian experience system optionally tracks per-tool reliability and distills cross-rollout strategy knowledge to progressively refine the adjudication process without parameter updates. We detail each component in the following sections.

\subsection{Knowledge-augmented Tool Planning}
\begin{figure*}[t!]
  \centering
   \includegraphics[width=1.\textwidth]{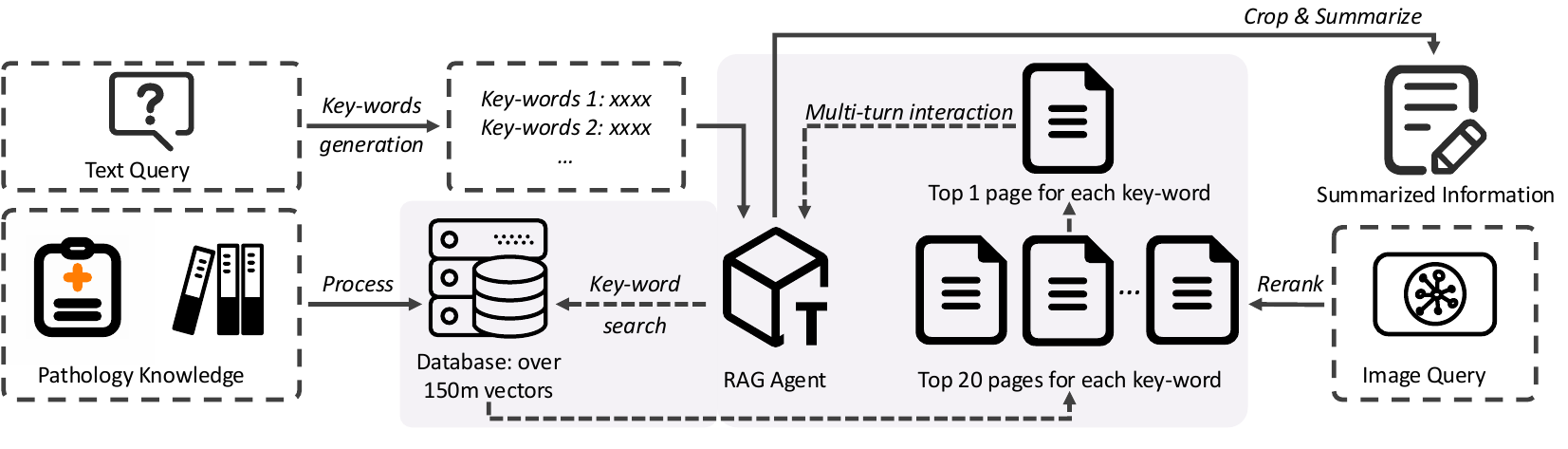}
   \caption{The RAG retriever pipeline. Per-candidate queries retrieve textbook pages from a Milvus database, which are read by the RAG agent and summarized into tool guidance.}
   \label{fig3}
\end{figure*}

To ground tool selection in domain knowledge rather than VLM intuition, PathoSage incorporates a retrieval-augmented knowledge module before tool execution begins, as illustrated in Fig.~\ref{fig3}.

\textbf{Pathology Knowledge Base.} We adopt the pathology textbook corpus from Patho-AgenticRAG~\cite{patho-agenticrag}, comprising over 200,000 curated pages from approximately 600 authoritative textbooks. Each page is embedded as an image--text pair using ColQwen2 \cite{fayssecolpali} into a shared vector space and indexed via HNSW \cite{malkov2018efficient} in Milvus \cite{wang2021milvus}, yielding a database $\mathcal{D}$ of over 150 million vectors that supports efficient joint text--image retrieval.

\textbf{Retrieval and Page Understanding.} Given the input query and image, we construct one or more keyword-based retrieval requests and retrieve the top-20 textbook pages per request from $\mathcal{D}$. Each top-ranked page is then processed by the VRAG agent $\mathcal{R}$ from Patho-AgenticRAG for multi-turn document understanding. The agent iteratively issues sub-queries, localizes relevant regions, and produces a final summary, autonomously terminating within at most three turns. This yields a structured knowledge summary $\mathcal{K} = \mathcal{R}(\mathcal{D}, Q, I)$ grounded in textbook evidence.

\textbf{Tool Guidance Generation.} The host VLM $\mathcal{V}$ receives $\mathcal{K}$ and generates a concise tool selection plan $\mathcal{G} = \mathcal{V}(\mathcal{K}, \mathcal{T}, Q, I)$, specifying which analysis dimensions (e.g., cellular morphology, tissue classification) are most relevant to the query. $\mathcal{G}$ is injected into the Phase A system prompt, and $\mathcal{K}$ is later provided to the SED independent assessment in Phase B as a reference for cross-checking tool conclusions.

\subsection{Phase A: ReAct Evidence Collection}
While pathology VLMs have achieved impressive performance, they remain susceptible to hallucinating morphological features that are not present in the image. Moreover, these models generate holistic textual descriptions but cannot perform quantitative analyses, such as cell counting and tissue-type distribution. In contrast, specialist models provide these capabilities, yet their outputs are heterogeneous in both format and scope.

To harness these complementary strengths, PathoSage includes a series of specialist tools spanning three analysis categories. Specifically, cell segmentation provides per-cell type counts and spatial distribution statistics. Patch classification maps image patches to tissue-type labels. Pathology-specialized VLM provides free-form diagnostic reasoning. This multi-dimensional analysis paradigm mirrors clinical pathology practice, where pathologists integrate morphological examination, quantitative biomarker scoring, and pattern-based judgment to reach a diagnosis \cite{pathchat, shaktah2025application}.

During Phase A, the host VLM $\mathcal{V}$ iteratively selects and invokes tools from $\mathcal{T}$ via function calling, guided by the tool plan $\mathcal{G}$. Each tool $t \in \mathcal{T}$ is paired with a structured operation description specifying its input requirements and output format, so that $\mathcal{V}$ can match the analytical needs identified in $\mathcal{G}$ to the appropriate tool capabilities. At each iteration $i$, $\mathcal{V}$ selects a tool $t_i$ and obtains its output, forming a reasoning step $\mathcal{P}_i$:
\begin{equation}
    P_i = \langle t_i, o_i \rangle = \langle t_i, t_i(I, Q) \rangle , \ t_i \in \mathcal{T},
\end{equation}
where $o_i$ represents the tool's structured result. Crucially, $\mathcal{V}$ is instructed to only collect evidence during this phase, which does not produce a final answer. This design reflects a deliberate separation of evidence gathering from evidence adjudication. By deferring judgment to Phase B's isolated contexts, we prevent the ReAct conversation history from anchoring the model's reasoning on early evidence. Phase A terminates when $\mathcal{V}$ invokes a dedicated termination function or reaches a maximum iteration count, yielding the collected evidence set $\mathcal{E} = \{\mathcal{P}_1, \mathcal{P}_2, \dots, \mathcal{P}_n\}$ that is forwarded to the structured deliberation procedure in Phase B.

\subsection{Phase B: Structured Evidence Deliberation}
When two or more tools are invoked, their outputs may conflict. For instance, a segmentation model may indicate predominantly inflammatory cells while a VQA model describes the tissue as neoplastic. To prevent the anchoring bias that arises from single-pass LLM fusion, we introduce Structured Evidence Deliberation (SED).

\textbf{Step 1: VLM Assessment.} The collected evidence $\mathcal{E}$ is first grouped by tool category. For each category, an independent VLM call containing only the original image $I$, query $Q$, the category's tool outputs, and the RAG knowledge $\mathcal{K}$ produces two semantic judgments: $a_i \in \{\text{agree},\; \text{uncertain},\; \text{disagree}\}$ and $r_i \in \{\text{high},\; \text{medium},\; \text{low}\}$, where $a_i$ indicates whether the VLM concurs with the conclusion of tool $t_i$ given its own visual understanding and textbook reference, and $r_i$ captures how relevant that conclusion is to the query. The assessor produces labels only, so that subsequent weighting is determined algorithmically rather than by VLM self-estimation \cite{xiongcan}.

\textbf{Step 2: Tool Conflict Analysis.} Given the assessments from step 1 and the tool reliability posteriors from the experience system, we compute a three-dimensional effective weight for each evidence source:
\begin{equation}
    w_i = \phi(r_i) \cdot \psi(a_i) \cdot \theta_i,
\end{equation}
where $\phi(\cdot)$ and $\psi(\cdot)$ are predefined numerical mappings (e.g., $\phi(\text{high}) = 1.0$, $\psi(\text{agree}) = 1.0$), and $\theta_i \in [0,1]$ is the historical reliability prior:
\begin{equation}
\theta_i = \frac{\alpha_i}{\alpha_i + \beta_i},
\end{equation}
which is initialized to 0.5 under the uninformative prior $\mathrm{Beta}(1,1)$ when no experience is available. The analyzer detects inter-category conflicts and produces a structured report $\mathcal{C}$ that ranks all evidence by $w_k$ and highlights disagreements.

\textbf{Step 3: Final Reasoning.} A final VLM call synthesizes the answer in yet another fresh context, receiving the original image $I$, query $Q$, the per-tool assessments, and the conflict report $\mathcal{C}$:
\begin{equation}
\mathcal{A} = \mathcal{V}\bigl(I,\; Q,\; \mathcal{E},\; \mathcal{C}\bigr).
\end{equation}
By isolating each step in a fresh LLM context, SED ensures that the final reasoning is informed by structured, pre-adjudicated evidence rather than raw, sequentially accumulated tool outputs.

\subsection{Experience System}
PathoSage supports a training-free experience system that progressively refines evidence adjudication across tasks. The system operates in two complementary layers: tracking per-tool reliability via Bayesian posterior updates, and distilling task-level strategy knowledge. 

\textbf{Layer 1: Tool Reliability Tracking.} We model each tool's reliability as a Beta-Bernoulli conjugate pair. For each tool $t_k \in \mathcal{T}$,  we maintain parameters $(\alpha_k, \beta_k)$. After each task with ground-truth feedback, we perform a continuous credit assignment that leverages the semantic assessment from SED Step 1. Concretely, let $R \in \{0, 1\}$ denote whether the final answer is correct, and let $s_k = \psi(a_k)$ and $v_k = \phi(r_k)$ be the mapped assessment and relevance scores. The update rule is:
\begin{equation}
(\alpha_k, \beta_k) \leftarrow \begin{cases}
    (\alpha_k + s_k \cdot v_k, \beta_k) &R = 1 \\
    (\alpha_k, \beta_k + (1 - s_k) \cdot v_k), &R = 0
\end{cases}
\end{equation}
A tool judged as ``agree'' with ``high'' relevance receives full credit when the task succeeds, while a tool judged ``disagree'' contributes minimally to either $\alpha$
or $\beta$, reflecting appropriate uncertainty about its role in the outcome. To generalize across visually similar inputs, we retrieve the top-$K$ nearest Beta records by image embedding similarity and aggregate them via distance-weighted averaging to form the prior for a new query. The posterior mean $\theta_k = \alpha_k / (\alpha_k + \beta_k)$ then feeds into the effective weight computation in SED Step 2.

\textbf{Layer 2: Cross Rollout Strategy Distillation.} While Layer 1 tracks statistical reliability, Layer 2 extracts symbolic strategy knowledge. During an exploration, we execute $N$ rollouts per query. Successful and failed rollouts are then compared to extract the semantic advantage, a natural-language description of what strategies led to success versus failure. Formally, given a group of rollouts $\{(y_j, R_j)\}_{j=1}^{N}$, where $y_j$ is the trajectory and $R_j$ is the binary outcome, we use the host VLM to introspect on the group and produce strategy updates via add, delete, or modify operations on a persistent strategy bank. The distilled strategies are injected into the Phase A system prompt, serving as a learned token prior that guides the VLM's behavior toward more effective tool calling.

\section{Experiments}
\setlength{\belowcaptionskip}{4pt}
\begin{table*}[t!]
\caption{Quantitative comparison of models on the PathMMU test set. The best result in each subset for general and pathology MLLMs is \textbf{in-bold}, and the second-best result is \underline{underlined}. Subscript \textcolor{green!60!black}{green numbers} indicate absolute performance gains relative to the host model (GPT-5.4).}
\renewcommand{\arraystretch}{1.0}
\renewcommand{\tabcolsep}{6pt}
\centering
\resizebox{\textwidth}{!}
{\begin{tabular}{lcccccccccccc}
\toprule[1pt]
\multirow{2}{*}{}        & \multicolumn{2}{c}{\textbf{Test Overall}} & \multicolumn{2}{c}{\textbf{PubMed}} & \multicolumn{2}{c}{\textbf{SocialPath}} & \multicolumn{2}{c}{\textbf{EduContent}} & \multicolumn{2}{c}{\textbf{Atlas}} & \multicolumn{2}{c}{\textbf{PathCLS}} \\  
                         & Tiny                & All                 & Tiny             & All              & Tiny               & All                & Tiny               & All                & Tiny             & All             & Tiny              & All              \\ \midrule[0.8pt]
\multicolumn{13}{c}{\textbf{General Multimodal LLMs}}                                                                                                                                                                                                              \\ \midrule[0.8pt]
InstructBLIP-FLAN-T5-XXL & 34.3                & 33.9                & 39.1             & 37.2             & 33.6               & 34.3               & 34.5               & 36.0               & 38.5             & 39.3            & 22.6              & 22.7             \\
LLaVA-1.5-13B            & 38.8                & 37.6                & 44.5             & 41.0             & 40.4               & 40.4               & 34.1               & 39.4               & 47.1             & 44.3            & 24.9              & 23.5             \\
LLaVA-Onevision-7B       & 36.9                & 34.4                & 37.7             & 36.4             & 35.7               & 38.4               & 47.1               & 38.3               & 38.9             & 38.4            & 20.3              & 20.4             \\
InternVL2.5-8B           & 50.1                & 48.6                & 55.9             & 53.0             & 57.8               & 54.4               & 50.6               & 50.6               & 51.4             & 50.8            & 29.4              & 31.1             \\
InternVL3-8B             & 55.4                & 52.9                & 57.7             & 55.9             & 60.6               & 57.0               & 54.9               & 52.9               & 58.2             & 56.2            & 42.9              & 41.0             \\
Qwen3-VL-8B-Instruct     & 55.9                    & 54.8                    & 61.9                 & 59.4                 & 60.6                   & 58.5                   & 55.3                   & 55.6                   & 54.8                 & 55.6                & 42.9                  & 42.6                 \\
Qwen3-VL-32B-Instruct    & 65.4                    & 61.9                    & 71.5                 & 64.2                 & 67.9                   & 65.5                   & 69.0                   & 65.4                   & 63.0                 & 66.8                & 50.3                  & 49.4                 \\
GPT-5.4-mini             & 70.1                    & 66.5                    & 73.0                 & 69.9                 & 72.9                   & 68.9                   & 79.6                   & 71.6                   & 69.2                 & 68.5                & 53.7                  & 50.4                 \\
GPT-5.4                  & 72.9                    & 70.8                    & 76.5                 & 74.2                 & 73.4                   & 71.2                   & 79.6                   & 75.4                   & 75.5                 & 75.1                & 53.7                  & 55.5                 \\
Gemini-3-Flash-Preview           & 78.1                    & 76.5                    & \underline{79.1}                 & \underline{78.8}                 & 74.3                   & \underline{74.5}                   & \underline{81.2}                   & 77.2                   & \underline{85.6}                 & \underline{83.8}                & 65.5                  & 66.1                 \\
Gemini-3-Pro-Preview             & \underline{78.6}                    & \textbf{77.1}                    & \textbf{79.4}                 & \textbf{79.0}                 & \underline{76.6}                   & \underline{74.5}                   & 80.8                   & \underline{77.5}                   & \textbf{86.1}                 & \textbf{85.4}                & 67.8                  & 69.2                 \\ \midrule[0.8pt]
\multicolumn{13}{c}{\textbf{Pathology Multimoda LLMs}}                                                                                                                                                                                                             \\ \midrule[0.8pt]
LLaVA-Med                & 25.3                & 26.2                & 28.5             & 27.7             & 28.9               & 27.3               & 22.7               & 27.2               & 22.6             & 30.7            & 22.6              & 20.3             \\
Quilt-LLaVA              & 45.6                & 41.5                & 47.3             & 42.6             & 46.4               & 46.6               & 51.8               & 45.3               & 46.2             & 42.7            & 32.2              & 29.2             \\
HuatuoGPT-7B             & 58.2                & 56.4                & 61.9             & 61.7             & 58.7               & 58.6               & 60.0               & 57.4               & 65.9             & 62.0            & 40.1              & 38.4             \\
PathGen-LLaVA            & 60.1                & 58.4                & 60.1             & 60.1            & 60.9               & 58.8               & 60.8               & 60.7               & 63.5             & 64.9            & 54.2              & 48.9             \\
CPath-Omni               & 72.4                & 72.2                & 74.0             & 69.9             & \underline{76.6}               & 71.8               & 69.8               & 70.6               & 65.9             & 70.6            & \textbf{75.7}              & \textbf{79.0}             \\
Patho-R1                 & 69.5                & 66.5                & 72.2             & 69.2             & 67.9               & 67.9               & 75.3               & 70.9               & 81.7             & 78.5            & 44.6              & 45.0             \\
Patho-AgenticRAG         & 73.2                & 70.5                & 72.2             & 71.0             & 74.7               & 72.9               & 76.5               & 73.8               & 79.3             & 78.8            & 57.2              & 55.2             \\
\textbf{\cellcolor{blue!10}PathoSage (GPT-5.4 as Host)}                & \textbf{\cellcolor{blue!10}79.6}\textcolor{green!60!black}{$_{\uparrow \textbf{6.7}}$}                    & \underline{\cellcolor{blue!10}76.8}\textcolor{green!60!black}{$_{\uparrow \textbf{6.0}}$}                    & \textbf {\cellcolor{blue!10}79.4}\textcolor{green!60!black}{$_{\uparrow \textbf{2.9}}$}                 & \cellcolor{blue!10}77.1\textcolor{green!60!black}{$_{\uparrow \textbf{2.9}}$}                 & \textbf{\cellcolor{blue!10}78.9}\textcolor{green!60!black}{$_{\uparrow \textbf{5.5}}$}                   & \textbf{\cellcolor{blue!10}76.5}\textcolor{green!60!black}{$_{\uparrow \textbf{5.3}}$}                   & \textbf{\cellcolor{blue!10}82.8}\textcolor{green!60!black}{$_{\uparrow \textbf{3.2}}$}                   & \textbf{\cellcolor{blue!10}79.2}\textcolor{green!60!black}{$_{\uparrow \textbf{3.8}}$}                   & \cellcolor{blue!10}83.7\textcolor{green!60!black}{$_{\uparrow \textbf{8.2}}$}                 & \cellcolor{blue!10}82.6\textcolor{green!60!black}{$_{\uparrow \textbf{7.5}}$}                 & \underline{\cellcolor{blue!10}71.2}\textcolor{green!60!black}{$_{\uparrow \textbf{17.5}}$}                  & \underline{\cellcolor{blue!10}71.0}\textcolor{green!60!black}{$_{\uparrow \textbf{15.5}}$}                 \\ \bottomrule[1pt]
\end{tabular}
}
\label{table1}
\end{table*}

\setlength{\belowcaptionskip}{4pt}
\begin{table*}[t!]
\caption{Quantitative comparison of models on Quilt-VQA, Path-VQA, MedXpert, and OmniMed. The best result in each subset for general and pathology MLLMs is \textbf{in-bold}, and the second-best result is \underline{underlined}.}
\renewcommand{\arraystretch}{1.2}
\renewcommand{\tabcolsep}{6pt}
\centering
\resizebox{0.6\textwidth}{!}
{
\begin{tabular}{lcccc}
\toprule[1pt]
                      & \multicolumn{2}{c}{\textbf{YorN}} & \textbf{MedXpert} & \textbf{OmniMed} \\ \cline{2-5} 
                      & Quilt       & Path       & Path     & Bright  \\ \midrule[1pt]
LLaVA-Onevision-7B    & 24.2        & 52.4       & 16.7     & 31.5    \\
InternVL2.5-8B        & 60.6        & 61.4       & 10.0     & 40.6    \\
InternVL3-8B          & 60.1        & 64.8       & 22.2     & 49.8    \\
Qwen3-VL-8B-Instruct  & 59.5        & 69.2       & 24.5     & 54.7    \\
Qwen3-VL-32B-Instruct & 58.9        & 63.9       & 22.2     & 61.6    \\
GPT-5.4-mini          & 65.9        & 68.8       & 46.7     & 56.9    \\
Gemini-3-Flash-Preview        & 74.1        & 74.1       & \textbf{80.0}     & 70.8    \\
Patho-R1              & 64.7        & 47.0       & 22.0     & 70.8    \\
Patho-AgenticRAG      & 75.8        & 80.3       & 60.0     & \textbf{90.1}    \\
\textbf{\cellcolor{blue!10}PathoSage (GPT-5.4 as Host)}            & \textbf{\cellcolor{blue!10}81.4}             & \textbf{\cellcolor{blue!10}83.2}           & \underline{\cellcolor{blue!10}71.1}         & \underline{\cellcolor{blue!10}88.3}        \\ \bottomrule[1pt]
\end{tabular}
}
\label{table2}
\end{table*}

We conduct comprehensive patch understanding evaluations across five diverse pathology Visual Question Answering (VQA) datasets, including PathMMU \cite{sun2024pathmmu}, Path-VQA \cite{he2020pathvqa}, Quilt-VQA \cite{quilt1m}, MedXpertQA \cite{zuo2025medxpertqa}, and OmniMedVQA \cite{hu2024omnimedvqa}. Detailed information of datasets and implementation are listed in Appendix.

\subsection{Quantitative Comparison}

\textit{\textbf{PathoSage significantly enhances the reasoning capability of its host model.}} As presented in Table \ref{table1}, PathoSage (using GPT-5.4 as the host) achieves an impressive 79.6\% on the PathMMU-test-tiny split and 76.8\% on the Test-All split. Notably, it substantially outperforms its own underlying reasoning engine, GPT-5.4, yielding an absolute improvement of 6.7\% on the Tiny split and 6.0\% on the All split. Furthermore, PathoSage remains highly competitive against Gemini-3-Pro, the strongest general MLLM evaluated, even surpassing it on the Test-Tiny overall score (79.6\% vs. 78.6\%) and dominating in specific subsets such as EduContext (82.8\% vs. 80.8\%) and PathCLS (71.2\% vs. 67.8\%). This superiority extends to the diverse diagnostic benchmarks in Table \ref{table2}, where PathoSage achieves 81.4\% on Quilt-VQA and 88.3\% on OmniMed. The performance advantage stems from the fundamental difference in reasoning paradigms. General MLLMs, despite their massive parameter counts, rely on single-pass generation. This "black-box" approach makes them susceptible to morphological hallucinations and anchoring biases when faced with complex, high-resolution pathology images. In contrast, PathoSage explicitly decouples evidence collection from adjudication. By integrating visual tools directly into the reasoning process and utilizing SED to weigh evidence, PathoSage grounds its diagnosis in verified features, thereby overcoming the inherent limitations of single MLLMs.

\textit{\textbf{PathoSage establishes a new state-of-the-art paradigm for domain-specific reasoning.}} Compared to Patho-AgenticRAG, which also employs retrieval and reasoning mechanisms, PathoSage achieves a significant lead (+6.7\% on PathMMU-test-tiny). For Yes/No questions requiring definitive morphological judgments, it achieves 81.4\% on Quilt-VQA and 83.2\% on Path-VQA, markedly outperforming previous pathology MLMMs. The results highlight the efficiency and robustness of our collaborative agentic design. Most existing pathology MLLMs require extensive domain-specific datasets and computationally expensive fine-tuning pipelines to acquire medical reasoning capabilities. In contrast, PathoSage is a training-free framework that achieves superior performance simply by orchestrating and collaborating existing specialized models. 

\subsection{Qualitative Analysis}
\begin{figure*}[t!]
  \centering
   \includegraphics[width=1.\textwidth]{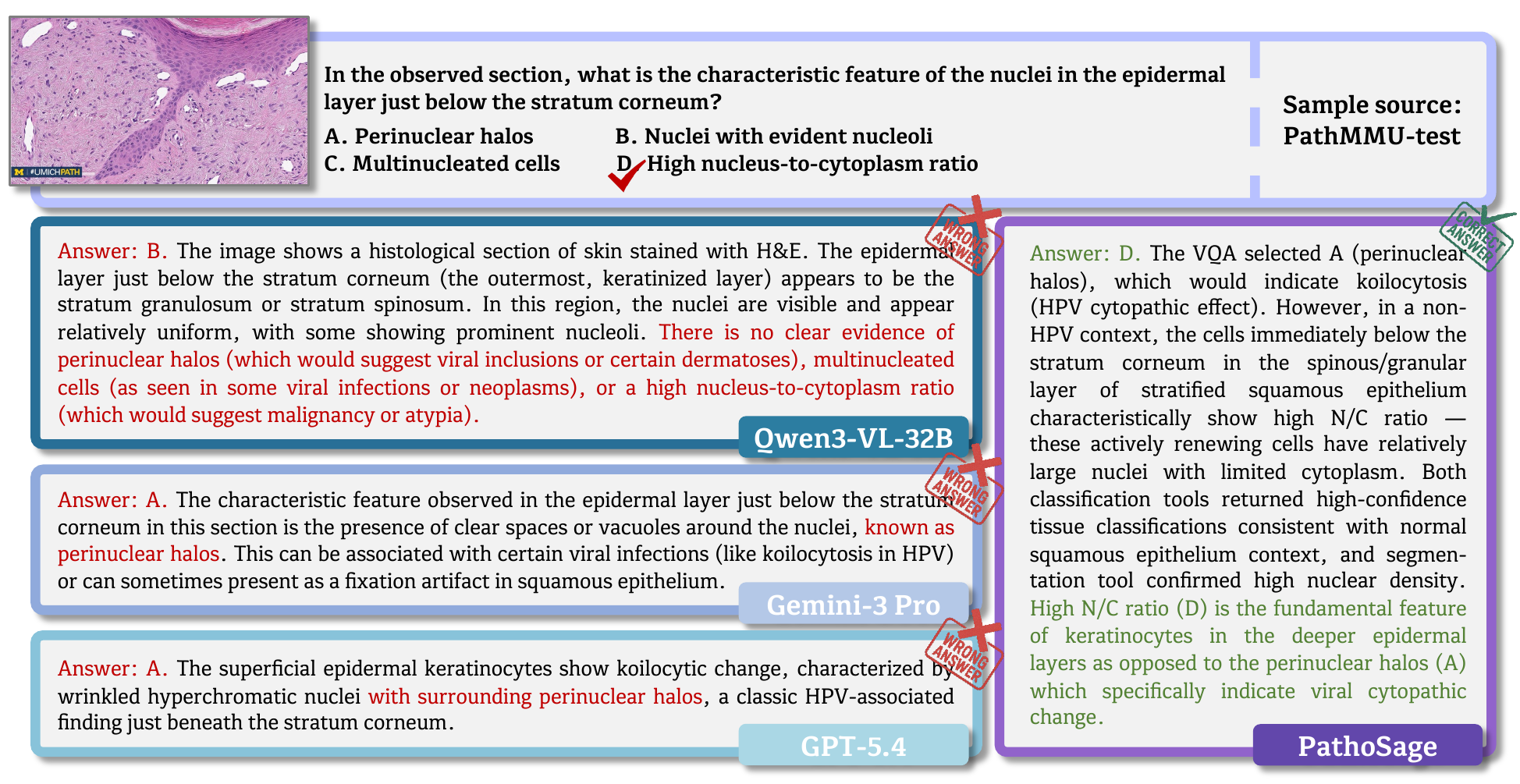}
   \caption{A representative example where PathoSage correctly identifies the answer, while three baseline MLLMs (Gemini-3-Pro, GPT-5.4, and Qwen2.5-32B) all fail on the same question.}
   \label{fig4}
\end{figure*}

Figure \ref{fig4} illustrates a representative VQA example from the PathMMU test set. This case highlights the vulnerability of MLLMs to visual deception. The baseline models, including the highly advanced Gemini-3 Pro and GPT-5.4, all fail by hallucinating "perinuclear halos" (Option A), a feature typically associated with viral infections (e.g., HPV) that is absent in this normal tissue section. Qwen3-VL-32B also misinterprets the visual context. In contrast, PathoSage correctly identifies the "high nucleus-to-cytoplasm ratio" (Option D). It achieves this not through a single-pass guess, but by orchestrating specialized tools: the classification tools confirm a normal squamous epithelium context, and the segmentation tool verifies high nuclear density. By grounding its reasoning in these verified tool outputs, PathoSage successfully avoids the hallucination trap. Additional qualitative examples, including detailed case studies on how SED explicitly resolves complex inter-tool conflicts by downweighting erroneous VQA suggestions, are provided in the Appendix.

\subsection{Ablation Study}
\begin{table*}[h!]
\caption{Ablation on key components, including RAG, Tool-box, SED, and Experience. The study is conducted on PathMMU-test-tiny. The best results are \textbf{in-bold}.}
\renewcommand{\arraystretch}{1.2}
\renewcommand{\tabcolsep}{6pt}
\centering
\resizebox{\textwidth}{!}
{
\begin{tabular}{cccccccccc}
\toprule[1pt]
\multicolumn{4}{c}{\textbf{Settings}}      & \multirow{2}{*}{\textbf{Test Overall}} & \multirow{2}{*}{\textbf{PubMed}} & \multirow{2}{*}{\textbf{SocialPath}} & \multirow{2}{*}{\textbf{EduContent}} & \multirow{2}{*}{\textbf{Atlas}} & \multirow{2}{*}{\textbf{PathCLS}} \\ \cline{1-4}
RAG & Tool-box & SED & Experience &                               &                         &                             &                            &                        &                          \\ \midrule[1pt]
     &           &      &             & 72.9                          & 76.5                    & 73.4                        & 79.6                       & 75.5                   & 53.7                     \\
\Checkmark   &           &      &             & 74.5                          & 76.9                    & 74.3                        & 80.0                       & 77.4                   & 59.3                     \\
     & \Checkmark        &      &             & 75.0                          & 76.9                    & 73.9                        & 81.2                       & 76.4                   & 62.7                     \\
     & \Checkmark        & \Checkmark   &             & 76.4                          & 77.9                    & 76.2                        & 80.4                       & 79.8                   & 65.5                     \\
\Checkmark   & \Checkmark        &      &             & 75.8                          & 77.2                    & 74.8                        & 81.2                       & 77.9                   & 64.4                     \\
\Checkmark   & \Checkmark        & \Checkmark   &             & 77.6                          & 78.3                    & 76.2                        & 81.6                       & 81.3                   & 68.4                     \\
\cellcolor{blue!10}\Checkmark   & \cellcolor{blue!10}\Checkmark        & \cellcolor{blue!10}\Checkmark   & \cellcolor{blue!10}\Checkmark          & \textbf{\cellcolor{blue!10}79.6}                          & \textbf{\cellcolor{blue!10}79.4}                    & \textbf{\cellcolor{blue!10}78.9}                        & \textbf{\cellcolor{blue!10}82.8}                       & \textbf{\cellcolor{blue!10}83.7}                   & \textbf{\cellcolor{blue!10}71.2}                     \\ \bottomrule[1pt]
\end{tabular}
}
\label{table3}
\end{table*}

\begin{figure*}[t!]
  \centering
   \includegraphics[width=1.\textwidth]{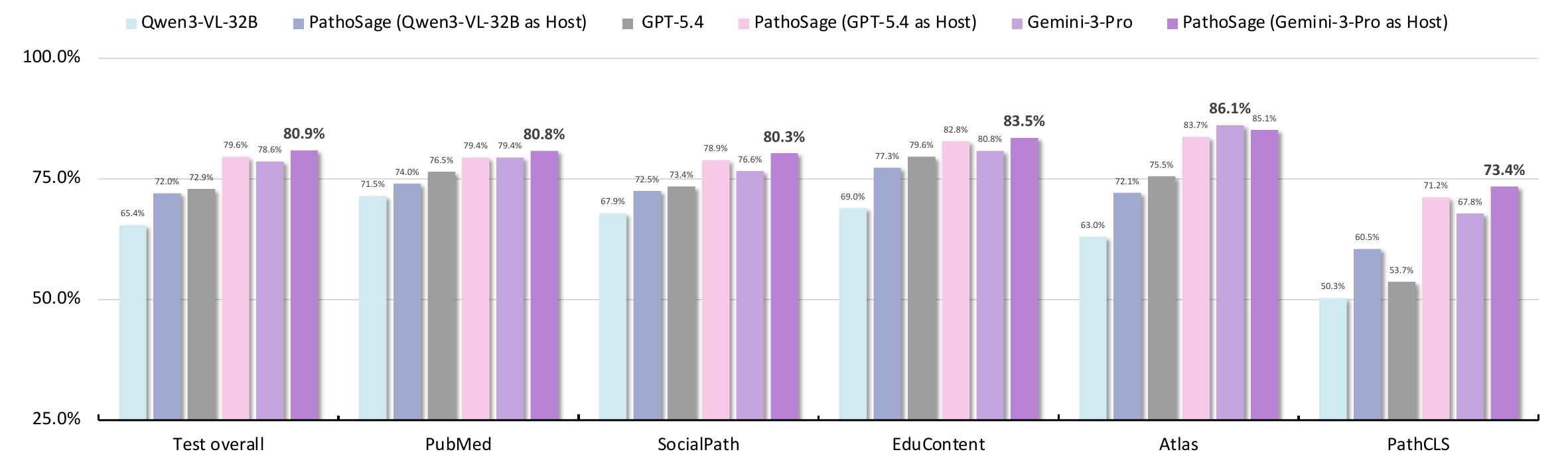}
   \caption{Ablation on the host VLM, including Qwen3-VL-32B, GPT-5.4, and Gemini-3-Pro. The study is conducted on PathMMU-test-tiny.}
   \label{fig5}
\end{figure*}

\textbf{Effectiveness of Key Components.} We ablate the core components of PathoSage on PathMMU-test-tiny. As shown in Table \ref{table3}, the host model (GPT-5.4) achieves an overall accuracy of 72.9\%. Introducing RAG alone improves performance to 74.5\% (+1.6\%), while introducing the Tool-box alone yields 75.0\% (+2.1\%), indicating that both external knowledge and specialized visual tools provide valuable diagnostic signals. However, simply combining them only reaches 75.8\%. This marginal gain (+0.8\% over Tool-box alone) suggests that naively aggregating heterogeneous evidence into a single context limits the utilization of all available information. When SED is applied to the Tool-box (without RAG), performance jumps to 76.4\% (+1.4\% over pure Tool-box). When SED is applied to both RAG and Tool-box, the accuracy increases to 77.6\% (+1.8\% over naive combination). Finally, incorporating the experience system further pushes accuracy to 79.6\% (+2.0\%), validating that modeling long-term tool reliability and establishing distance-weighted priors are crucial for resolving complex inter-tool conflicts.

\textbf{Generalizability Across Host VLMs.} To verify that the performance gains of PathoSage are not restricted to a specific model, we evaluate our framework using three host VLMs. As illustrated in Figure \ref{fig5}, PathoSage consistently enhances the accuracy across all tested hosts. Specifically, PathoSage improves the overall accuracy of Qwen3-VL-32B from 65.4\% to 72.0\% (+6.6\%), GPT-5.4 from 72.9\% to 79.6\% (+6.7\%), and Gemini-3-Pro from 78.6\% to 80.9\% (+2.3\%). Notably, when equipped with Gemini-3-Pro, PathoSage achieves the highest performance across almost all subsets. Furthermore, when using the open-source Qwen3-VL-32B, PathoSage (72.0\%) performs competitively with the single GPT-5.4 model (72.9\%). These consistent improvements confirm that our decoupled evidence adjudication and experience-aware routing mechanisms provide a robust, model-agnostic paradigm for advancing pathology AI.

\section{Conclusion}
\textbf{Broader Impact.} This paper introduces PathoSage, an agentic framework designed to deliver reliable pathology diagnoses by explicitly decoupling evidence collection from final adjudication. Addressing the limitations of VLMs and naive agentic workflows that suffer from context contamination, PathoSage incorporates SED to algorithmically weigh tool outputs and an experience system to model long-term tool reliability. By bridging the gap between opaque AI predictions and the rigorous, evidence-based diagnostic processes of human pathologists, our method represents a significant step toward trustworthy and clinically translatable AI in computational pathology.

\textbf{Limitations.} The proposed framework's performance inherently depends on the availability and accuracy of specialized visual tools, which remain limited for rare disease subtypes. In addition, the final diagnostic synthesis still relies on the host VLM, which is prone to inherent inconsistencies and hallucinations if all supporting tools provide erroneous evidence. Addressing these limitations through end-to-end tool optimization and more granular error taxonomies will be essential to further improve the clinical impact of computer-aided diagnosis.

\begin{ack}
This work was supported in part by the Natural Science Foundation of Sichuan Province under Grant 2026NSFSC1491, the National Natural Science Fundation of China (Grant No. 62303338, No. 62427820), the Fundamental Research Funds for the Central Universities under grant Sichuan University YJ202285, the Sichuan Science and Technology Program under Grant 2025ZDZX0125, the Science Fund for Creative Research Groups of Sichuan Province Natural Science Foundation under Grant 2024NSFTD0035.
\end{ack}


\bibliographystyle{unsrt} 
\bibliography{neurips_2026}

\newpage
\setcounter{figure}{0} 
\renewcommand{\thefigure}{\thesection.\arabic{figure}}
\setcounter{table}{0}
\renewcommand{\thetable}{\thesection.\arabic{table}}
\appendix

\section{Additional Experiments and Discussion}

\subsection{Experience Accumulation on PathMMU-val}

\begin{table*}[h!]
\caption{Quantitative comparison of models on the PathMMU val set. The best result in each subset for general and pathology MLLMs is \textbf{in-bold}, and the second-best result is \underline{underlined}.}
\renewcommand{\arraystretch}{1.0}
\renewcommand{\tabcolsep}{6pt}
\centering
\resizebox{\textwidth}{!}
{
\begin{tabular}{lcccccc}
\toprule[1pt]
\multicolumn{1}{c}{\multirow{2}{*}{}} & \multicolumn{6}{c}{\textbf{PathMMU-val}}                              \\ \cline{2-7} 
\multicolumn{1}{c}{}                  & Overall & PubMed & SocialPath & EduContent & Atlas & PathCLS \\ \midrule[0.8pt]
\multicolumn{7}{c}{\textbf{General Multimodal LLMs}}                                                          \\ \midrule[0.8pt]
LLaVA-Onevision-7B                    & 23.4    & 30.0   & 23.3       & 19.2       & 20.0  & 16.7    \\
InternVL2.5-8B                        & 47.8    & 47.6   & 54.0       & 52.1       & 46.3  & 33.3    \\
InternVL3-8B                          & 49.8    & 52.4   & 54.0       & 45.9       & 52.5  & 40.6    \\
Qwen3-VL-8B-Instruct                  & 53.5    & 53.6   & 56.7       & 54.1       & 57.5  & 43.8    \\
Qwen3-VL-32B-Instruct                 & 59.7        & 61.4       & 61.3           & 61.6           & 71.2      & 40.6        \\
GPT-5.4-mini                          & 61.8    & 68.2   & 62.0       & 56.9       & 70.0  & 46.9    \\
GPT-5.4                               & 65.4    & 71.2   & 64.7       & 58.9       & 70.0  & 58.3    \\
Gemini-3-Flash-Preview                & 73.2    & 75.5   & 67.3       & \underline{74.7}       & \underline{81.3}  & 67.7    \\
Gemini-3-Pro-Preview                  & \underline{74.5}    & \underline{81.5}   & 66.7       & 69.2       & 80.0  & \textbf{72.9}    \\ \midrule[0.8pt]
\multicolumn{7}{c}{\textbf{Pathology Multimodal LLMs}}                                                        \\ \midrule[0.8pt]
LLaVA-Med                             & 17.9    & 18.9   & 18.0       & 20.6       & 22.5  & 7.3     \\
Quilt-LLaVA                           & 33.1    & 34.3   & 34.7       & 32.9       & 45.0  & 17.7    \\
HuatuoGPT                             & 54.6    & 55.4   & 60.7       & 54.1       & 61.3  & 38.5    \\
PathGen-LLaVA                         & 58.2    & 59.7   & 53.3       & 61.0       & 67.5  & 50.0    \\
Patho-R1                              & 63.0    & 64.0   & 64.7       & 63.0       & \textbf{82.5}  & 41.7    \\
PathoSage (GPT-5.4 as Host, Pass@1)   & 74.0        & 80.7       & \underline{72.7}           & 69.9           & 75.0      & 65.6        \\
\textbf{\cellcolor{blue!10}PathoSage (GPT-5.4 as Host, Pass@4)}   & \textbf{\cellcolor{blue!10}80.1}    & \textbf{\cellcolor{blue!10}86.3}   & \textbf{\cellcolor{blue!10}76.0}       & \textbf{\cellcolor{blue!10}78.8}       & \textbf{\cellcolor{blue!10}82.5}  & \underline{\cellcolor{blue!10}71.9}    \\ \bottomrule[1pt]
\end{tabular}
}
\label{table_val}
\end{table*}

To ensure a rigorous evaluation and prevent data leakage during the testing phase, PathoSage's Beta-Bernoulli experience system is exclusively accumulated on the validation set of PathMMU (PathMMU-val). This initial exploration phase serves as the foundation for modeling long-term tool reliability and extracting task-level strategies. Table \ref{table_val} presents the quantitative performance of various general and pathology-specific MLLMs on the PathMMU val set, alongside the performance of PathoSage during this critical accumulation phase.

During the experience accumulation process, PathoSage operates in \textit{exploration mode}. We evaluate its performance under two distinct settings:
\begin{itemize}
    \item \textbf{Pass@1:} The agent performs a single reasoning trajectory without utilizing any historical experience priors ($\theta_k = 0.5$). This serves as the baseline performance of our SED mechanism acting alone.
    \item \textbf{Pass@4:} The agent executes $N=4$ independent rollouts per query. The system aggregates the results from these multiple trajectories to extract successful strategies and update the Beta-Bernoulli reliability parameters.
\end{itemize}

As shown in Table \ref{table_val}, PathoSage (Pass@1), utilizing GPT-5.4 as the host model, achieves a strong overall accuracy of 74.0\% on the validation set. This represents a substantial absolute improvement of \textbf{+8.6\%} over the bare GPT-5.4 (65.4\%) and performs competitively with the strongest general-purpose baseline, Gemini-3-Pro (74.5\%). This significant margin confirms that explicitly decoupling evidence collection from adjudication via SED provides a highly robust reasoning foundation, even before any historical experience is accumulated.

Crucially, when operating in exploration mode (Pass@4), PathoSage's performance surges to an impressive overall accuracy of \textbf{80.1\%}, establishing a new state-of-the-art across nearly all sub-categories (e.g., 86.3\% on PubMed and 78.8\% on EduContent). The multiple rollouts allow the system to explore diverse tool combinations and reasoning paths, successfully navigating complex cases where a single-pass generation might fail. 

The successful trajectories identified during this Pass@4 exploration are subsequently harvested to populate the experience database. This process yields high-quality pseudo-labels for updating the Beta-Bernoulli reliability parameters ($\alpha_k, \beta_k$) and extracts valuable strategies for tool utilization. The accumulated experience from this validation phase is then frozen and utilized to establish priors during the final evaluation on the PathMMU test set.

\subsection{Analysis of Experience Database}

\begin{figure*}[h!]
  \centering
   \includegraphics[width=1.\textwidth]{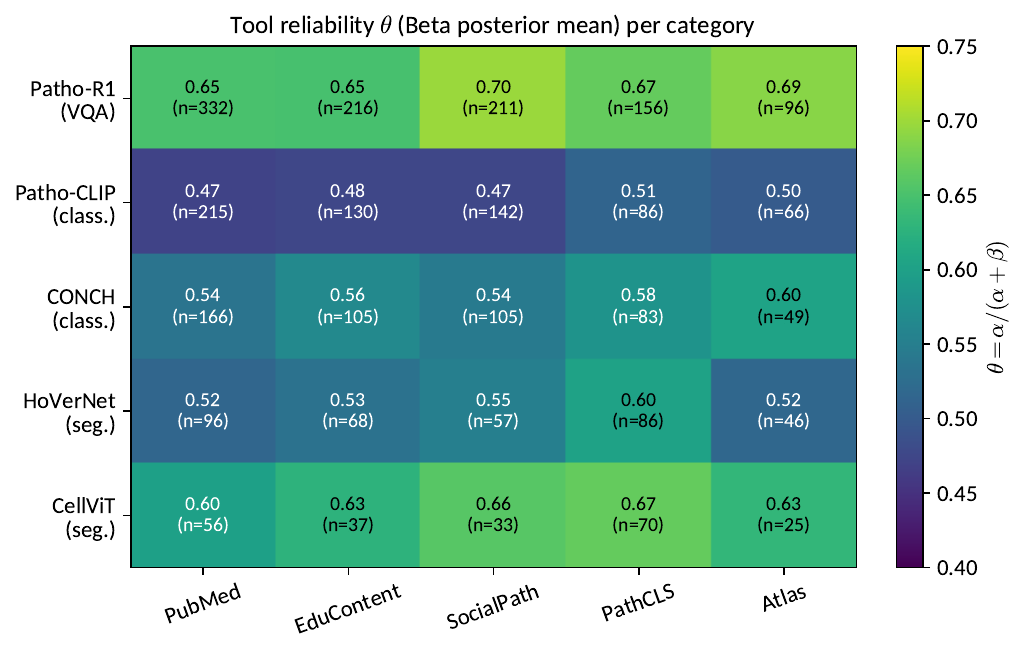}
   \caption{Beta posterior mean $\theta$ for each tool per category and number of labeled samples.}
   \label{layer_1}
\end{figure*}

To understand what PathoSage learns during the initial exploration phase, we conduct an in-depth analysis of the accumulated experience database on the PathMMU val set. We examine both the statistical tool reliability tracked by Layer 1 (Beta-Bernoulli updates) and the symbolic strategy knowledge distilled by Layer 2 (Cross-Rollout Critique).

\textbf{Layer 1: Tool Reliability Profiling.} Figure \ref{layer_1} illustrates the posterior mean $\theta_k = \alpha_k / (\alpha_k + \beta_k)$ for each specialized tool across different PathMMU sub-categories. This metric represents the system's learned trust in a specific tool for a given domain. Several key observations emerge:
\begin{itemize}
    \item \textbf{Domain-Specific Competence:} Tool reliability is highly heterogeneous. For instance, the VQA tool (Patho-R1) maintains consistently high reliability ($\theta \approx 0.65-0.70$) across all categories, reflecting its strong general visual reasoning capabilities. Conversely, zero-shot classifiers (Patho-CLIP and CONCH) exhibit moderate reliability ($\theta \approx 0.47-0.60$), indicating that while they provide useful signals, their raw outputs should not unconditionally override other evidence without deliberation.

     \item \textbf{Sensitivity to Model Architectures:} Within the segmentation category, CellViT consistently achieves higher reliability scores ($\theta \approx 0.60-0.67$) compared to HoVerNet ($\theta \approx 0.52-0.60$). This suggests that the experience system successfully captures the underlying performance differences between tool architectures, naturally learning to prioritize the more robust CellViT model when resolving conflicts in downstream tasks.
\end{itemize}

These learned $\theta$ values validate the necessity of our distance-weighted priors: rather than treating all tools with static, uniform trust, PathoSage dynamically calibrates its reliance based on the specific tissue microenvironment and historical tool performance.

\begin{figure*}[h!]
  \centering
   \includegraphics[width=1.\textwidth]{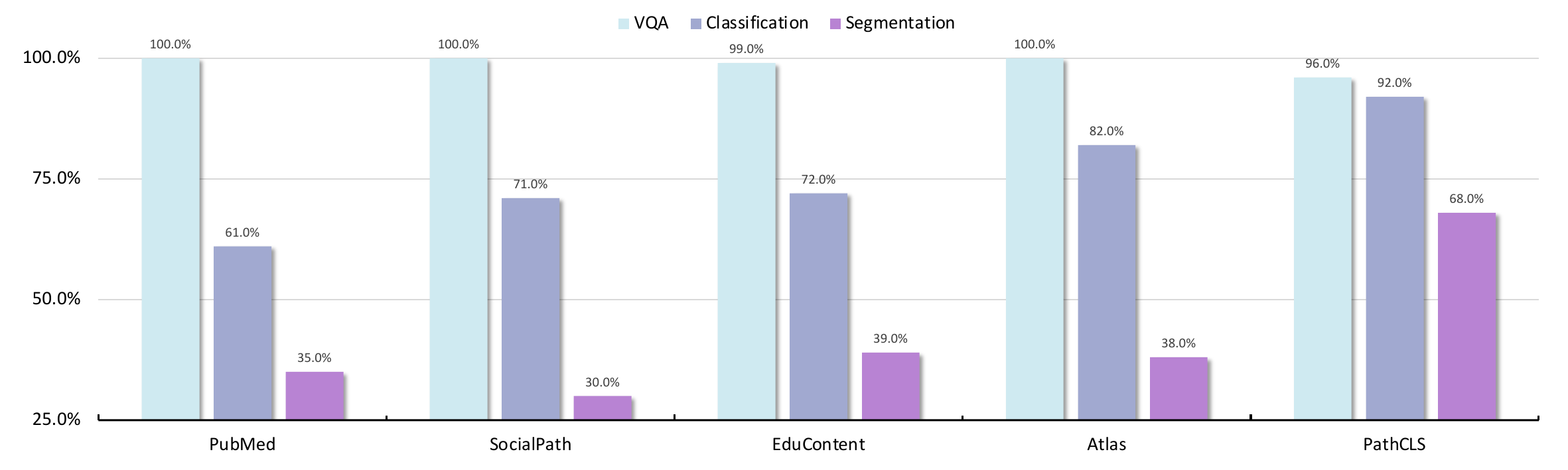}
   \caption{The percentage of tool categories marked as “critical” under each dataset category.}
   \label{layer_2}
\end{figure*}

\textbf{Layer 2: Cross-Rollout Critique and Tool Criticality.} Beyond statistical reliability, Layer 2 of the experience system distills task-level strategies by comparing successful and failed rollouts. During the exploration phase (4 rollouts per query), the system identifies \textit{critical tools}, which are defined as tool categories that consistently appear in successful trajectories and play a pivotal role in the reasoning chain, but are absent or misused in failed ones. 

Figure \ref{layer_2} presents the proportion of samples within each sub-category where a specific tool type was flagged as "critical". The distribution reveals a clear hierarchical strategy learned by the system:
\begin{itemize}
    \item \textbf{VQA as the Primary Anchor:} The VQA module is identified as critical in nearly 100\% of cases across all subsets. This indicates that the experience database recognizes VQA as the "workhorse" for PathMMU tasks, essential for interpreting complex, open-ended morphological queries.
    \item \textbf{Classification as a Secondary Validator:} Patch classification is deemed critical in 61.0\% to 92.0\% of cases, depending on the subset. It serves as a crucial auxiliary signal, particularly in subsets like PathCLS (92.0\%) and Atlas (82.0\%), where tissue-level identification is often required to ground the VQA's narrative.
    \item \textbf{Segmentation for Specialized Contexts:} Segmentation tools are marked as critical in a minority of cases (30.0\% to 68.0\%). Notably, its criticality peaks in the PathCLS subset (68.0\%), which predominantly consists of single H\&E patches where fine-grained cellular composition (e.g., nuclear density, cell counting) is decisive for the final diagnosis.
\end{itemize}
This analysis demonstrates that PathoSage does not merely learn to call all available tools blindly. Instead, it successfully distills a nuanced, pathology-aware strategy: anchoring on VQA for general reasoning, cross-validating with classifiers for tissue context, and selectively invoking segmentation for fine-grained cellular tasks.

\subsection{Examples of How PathoSage Handles Tool Evidence}

While introducing multiple tools provides richer context, it inevitably leads to conflicting evidence and invites uncritical agreement. PathoSage's SED mechanism is designed to handle both regimes explicitly.

\begin{figure*}[h!]
  \centering
   \includegraphics[width=1.\textwidth]{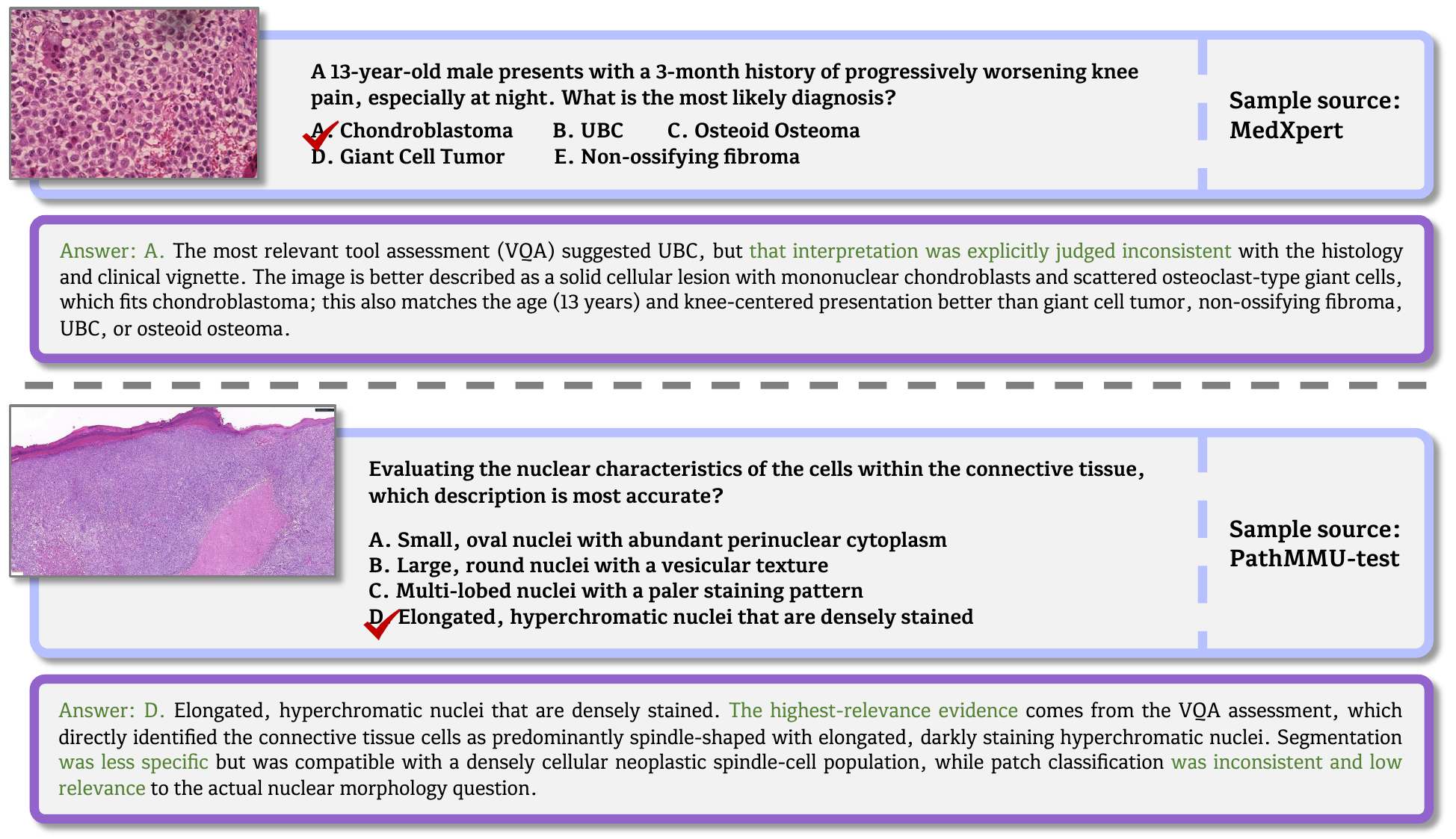}
   \caption{Two examples illustrating how PathoSage resolves tool conflicts. One indicates that an incorrect VQA suggestion is overridden; the other indicates that irrelevant classification outputs are downweighted in favor of high-relevance VQA evidence.}
   \label{tool_conflict_1}
\end{figure*}

Figure \ref{tool_conflict_1} demonstrates how PathoSage's SED mechanism explicitly adjudicates conflicts where a non-VQA evidence stream is at odds with the higher-relevance signal. In the first scenario (top, MedXpert), the VQA tool incorrectly suggested "UBC" for a bone lesion. A naive agentic system would likely suffer from anchoring bias and adopt this suggestion. However, PathoSage's SED independently evaluated the VQA output against the clinical vignette and histological features, explicitly judging the VQA interpretation as inconsistent. By downweighting this erroneous evidence, the system correctly concluded the diagnosis was Chondroblastoma (Option A). In the second scenario (bottom, PathMMU-test) regarding nuclear characteristics, the patch classification tool produced an inconsistent and low-relevance output, while the segmentation tool was compatible but less specific. SED correctly identified that the VQA assessment provided the highest-relevance evidence for this specific morphological query. By algorithmically prioritizing the VQA output and disregarding the irrelevant classifier, PathoSage accurately selected Option D.

\begin{figure*}[h!]
  \centering
   \includegraphics[width=1.\textwidth]{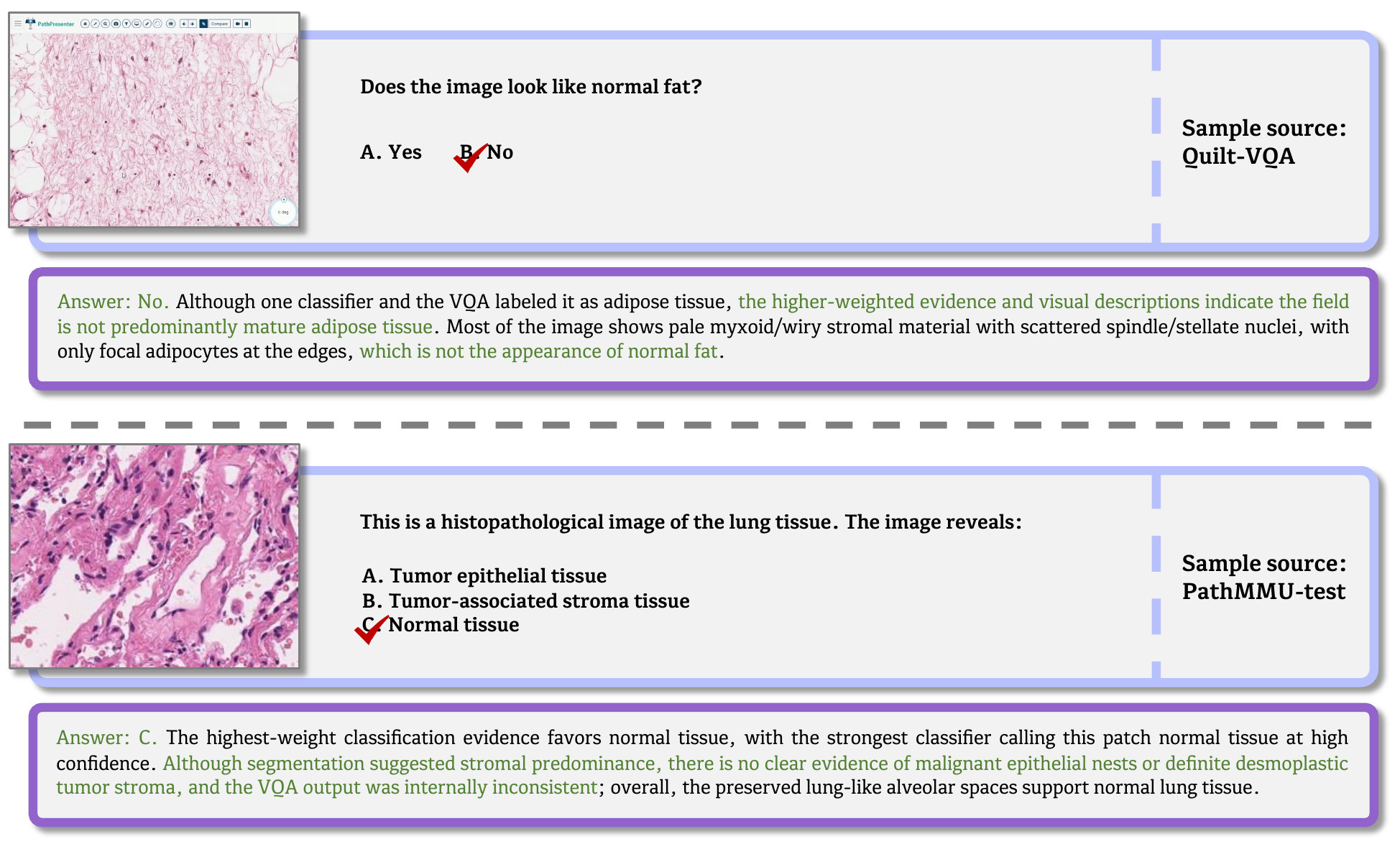}
   \caption{Two additional examples illustrating how PathoSage resolves tool conflicts. In both cases, SED detects inconsistencies in the VQA output and prioritizes high-confidence classifier evidence to recover the correct answer.}
   \label{tool_conflict_2}
\end{figure*}

Figure \ref{tool_conflict_2} extends this analysis to the harder case in which the VQA tool itself supplies the wrong interpretation. In the first example (top, Quilt-VQA), the VQA tool described the field as resembling normal fat, which would have led to an incorrect "Yes" answer. SED detected that the VQA reasoning was self-contradictory, and instead trusted the higher-weighted classifier evidence identifying the dominant component as myxoid stromal material, yielding the correct answer of "No." In the second example (bottom, PathMMU-test), the VQA tool selected Option B (tumor-associated stroma) for an H\&E lung patch, but the highest-confidence classifier called the tissue normal, and SED noted that the VQA narrative lacked decisive markers of malignant epithelial nests or desmoplastic stroma. SED therefore overrode the VQA suggestion and committed to Option C (normal tissue). These cases show that PathoSage's deliberation process is not biased toward any single tool.

\begin{figure*}[h!]
  \centering
   \includegraphics[width=1.\textwidth]{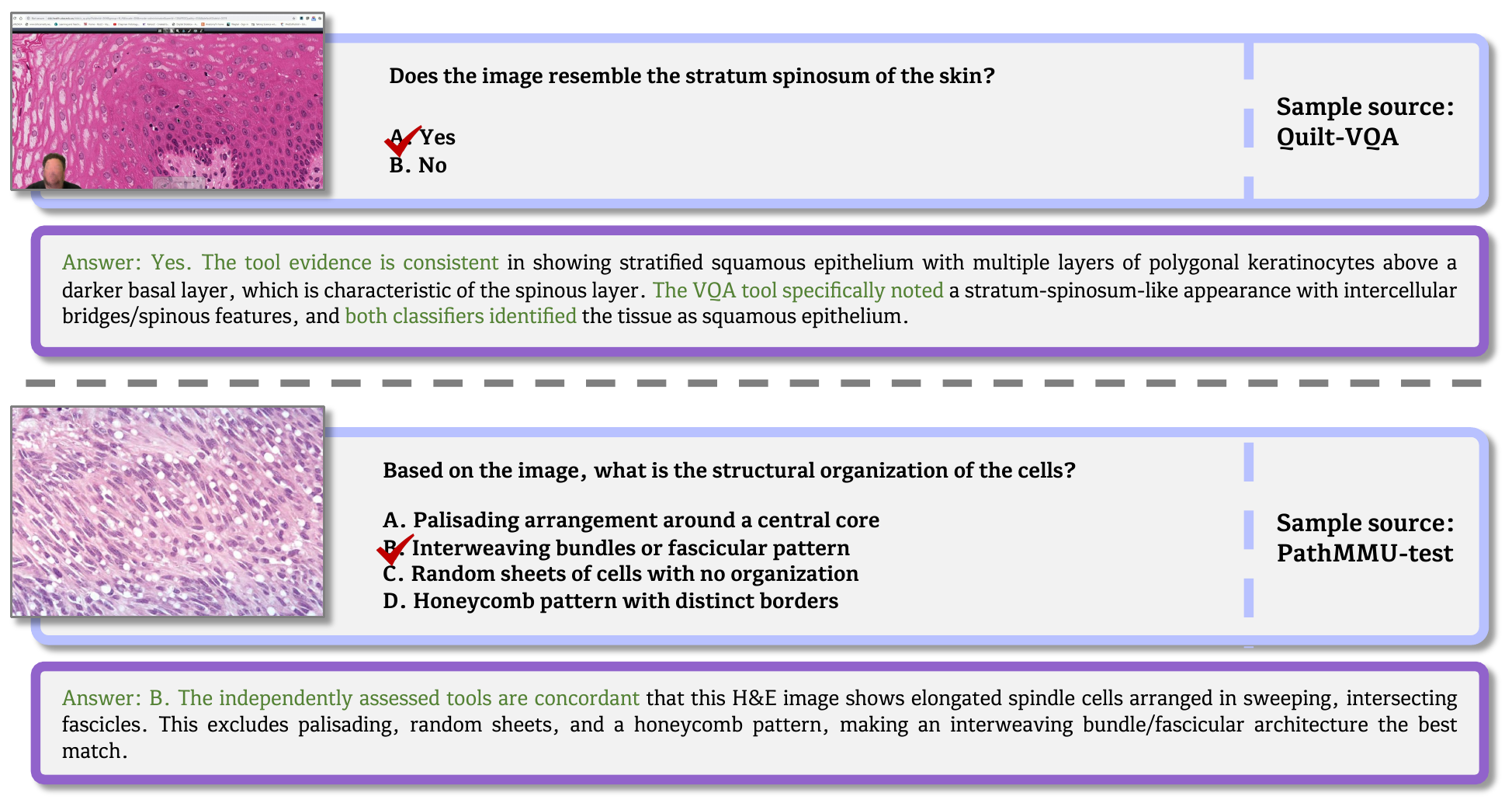}
   \caption{Two examples in which the independently invoked tools converge on consistent evidence. PathoSage routes such concordant signals through a streamlined deliberation path and commits to the correct answer with high confidence.}
   \label{tool_agree}
\end{figure*}

Figure \ref{tool_agree} illustrates the complementary regime in which the invoked tools converge on consistent evidence. In the first example (top, Quilt-VQA), the VQA tool described stratum-spinosum-like polygonal keratinocytes with intercellular bridges, and both classifiers independently labelled the tissue as squamous epithelium, jointly supporting the answer "Yes." In the second example (bottom, PathMMU-test), the VQA tool, the patch classifier, and the cell segmentation tool all reported elongated spindle cells arranged in sweeping fascicles, unambiguously matching Option B (interweaving bundle/fascicular pattern). Under such concordant evidence, SED routes the deliberation through a streamlined consolidation path, allowing PathoSage to commit to the correct answer with high confidence and minimal additional reasoning. Together, Figures \ref{tool_conflict_1}–\ref{tool_agree} demonstrate that PathoSage's SED mechanism delivers a consistent treatment of evidence across the full spectrum from open conflict to full consensus.

\subsection{Failure Case Analysis}
\begin{figure*}[h!]
  \centering
   \includegraphics[width=1.\textwidth]{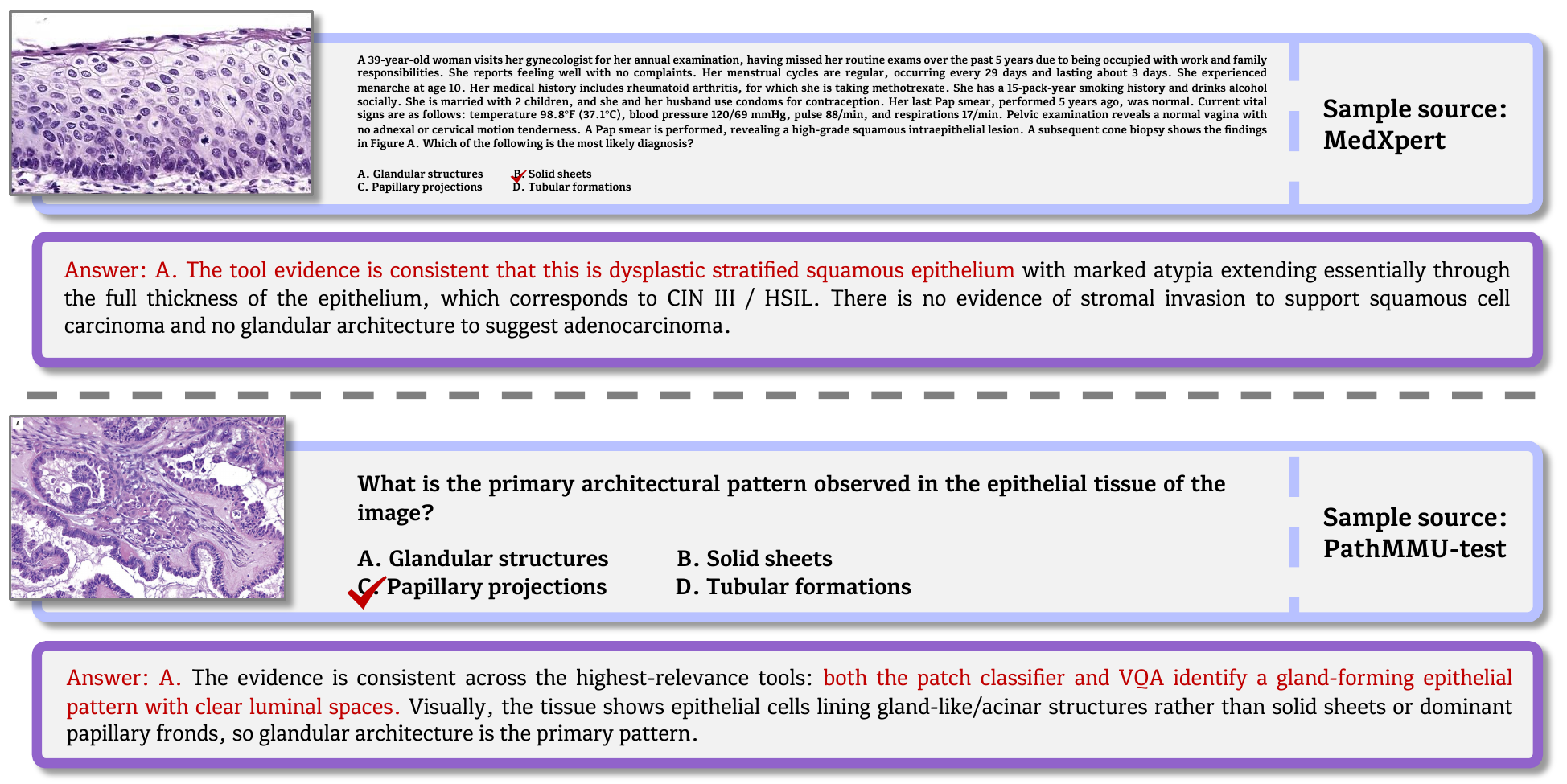}
   \caption{Two failure cases in which all invoked tools converge on the same, but ultimately incorrect interpretation.}
   \label{failure_case_1}
\end{figure*}

\begin{figure*}[h!]
  \centering
   \includegraphics[width=1.\textwidth]{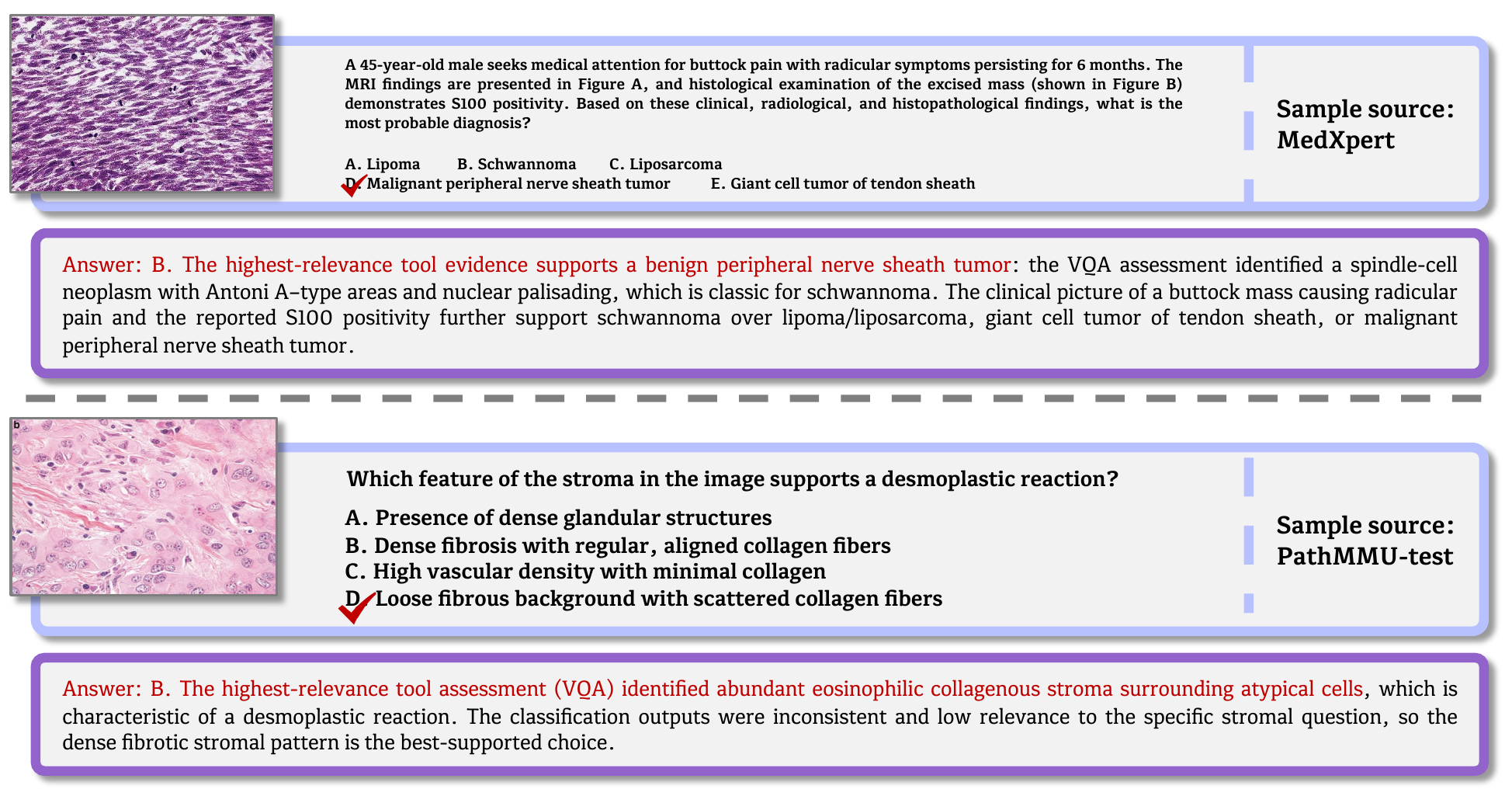}
   \caption{Two failure cases in which a confident but flawed VQA narrative dominates the deliberation.}
   \label{failure_case_2}
\end{figure*}

Despite the robustness demonstrated above, PathoSage still exhibits two characteristic failure modes, illustrated in Figures \ref{failure_case_1} and \ref{failure_case_2}.

Figure \ref{failure_case_1} shows cases of collective tool failure, in which all invoked tools converge on the same incorrect interpretation. In the first example (top, MedXpert), all four tools concurred on CIN III, although the ground-truth diagnosis was CIN II, reflecting a one-grade overcall of dysplasia severity. In the second example (bottom, PathMMU-test), the patch classifier, the VQA tool, and the cell segmentation tool unanimously identified a "glandular" architectural pattern, whereas the ground-truth label was papillary projections.  Because SED's central premise is that disagreement between independent evidence streams flags potential errors, this premise breaks down when the streams share the same systematic bias; no amount of additional deliberation within the existing tool ensemble can recover the correct answer.

Figure \ref{failure_case_2} illustrates a second failure mode in which a high-relevance VQA narrative is itself misleading and dominates the deliberation.  In the first example (top, MedXpert), a clinical case of an S100-positive spindle-cell tumor with radicular symptoms, the VQA tool fixated on Antoni A-type nuclear palisading and selected schwannoma (option B), whereas the ground truth was malignant peripheral nerve sheath tumor (option D). The disagreeing classifier outputs carried only medium relevance and were down-weighted. In the second example (bottom, PathMMU-test), the VQA tool described "abundant eosinophilic collagenous stroma surrounding atypical cells" and selected option B (dense, aligned collagen fibers); however, the ground truth was option D (loose fibrous background with scattered collagen fibers). The patch classifiers were inconsistent and were assigned low relevance, so SED's relevance-weighted aggregation could not counterbalance the confident but incorrect VQA assertion. These cases reveal that when the most diagnostic tool produces a self-consistent but erroneous narrative, lower-relevance corroborating tools may fail to generate sufficient counterweight to override it.

It is worth noting that these residual failures do not stem from a flaw in the SED mechanism itself, but rather expose the limits of the underlying tool ensemble. SED is, by construction, an aggregator of independent evidence: when the available evidence streams share the same systematic bias (Figure \ref{failure_case_1}), or when the most diagnostic tool produces a confident but erroneous narrative that no higher-relevance counter-evidence is available to challenge (Figure \ref{failure_case_2}), no purely aggregation-level rule can recover the correct answer. Redesigning SED to overrule a unanimous high-relevance signal on the basis of weaker disagreeing evidence would reintroduce the very anchoring and over-correction biases that SED was designed to suppress, and would degrade performance on the much larger set of cases where the consensus is in fact correct. The principled remedy is therefore not to retune the deliberation logic, but to enrich the evidence pool itself. We regard this as a natural direction for future extensions of PathoSage rather than a limitation of the current framework.

\clearpage
\section{Implementation Setup}
\subsection{Detailed Dataset Descriptions}

We conduct comprehensive patch understanding evaluations across five diverse pathology Visual Question Answering (VQA) datasets, encompassing both binary (Yes/No) and multiple-choice reasoning tasks. These datasets are selected to assess different dimensions of the agent's capabilities, ranging from definitive morphological judgments to complex differential diagnoses. The detailed statistics and configurations for each dataset are as follows:

\textbf{PathMMU.}
PathMMU \cite{sun2024pathmmu} serves as our primary evaluation suite due to its extensive coverage of diverse tissue types, clinical scenarios, and expert-validated annotations. Following its official protocol, we report results on both the full test set and a representative \textit{test-tiny} split, allowing for a rigorous assessment of PathoSage's ability to integrate heterogeneous tool evidence in high-resolution patch analysis. 
\begin{itemize}
    \item \textbf{Test-Tiny Split:} Comprises a total of 1,139 questions, distributed across five distinct sub-categories: Atlas (208), EduContent (255), PathCLS (177), PubMed (281), and SocialPath (218).
    \item \textbf{Test-All Split:} Scales up to 8,454 questions, distributed as: Atlas (799), EduContent (1,683), PathCLS (1,632), PubMed (2,787), and SocialPath (1,553).
\end{itemize}

\textbf{Multiple-Choice Diagnostic Benchmarks.} 
For more complex diagnostic reasoning and differential analysis, we employ multiple-choice questions sourced from MedXpertQA \cite{zuo2025medxpertqa} and OmniMedVQA \cite{hu2024omnimedvqa}. 
\begin{itemize}
    \item \textbf{MedXpertQA:} We filter the original dataset to extract 90 highly relevant pathology examples that require expert-level medical knowledge.
    \item \textbf{OmniMedVQA:} We utilize the BRIGHT Challenge subset, which consists of 890 cases focusing on challenging diagnostic reasoning across medical specialties.
\end{itemize}

\textbf{Binary Morphological Benchmarks.}
To assess the model's capability in making definitive morphological judgments, we utilize Yes/No (YorN) questions selected from the test splits of Path-VQA \cite{he2020pathvqa} and Quilt-VQA \cite{quilt1m}. These tasks require precise identification of specific pathological features (e.g., the presence of necrosis, specific cellular arrangements, or staining characteristics). We collect closed-ended questions from their respective test splits, resulting in 3,362 questions for Path-VQA and 343 questions for Quilt-VQA.

\subsection{Implementation Details and Compute Resources}
Throughout the PathoSage workflow, we employ GPT-5.4 \cite{GPT-5.4} as the host VLM for our main experiments. Although Gemini-3-Pro \cite{Gemini3pro} demonstrated marginally superior performance in preliminary tests, we selected GPT-5.4 to achieve an optimal trade-off between diagnostic accuracy and inference efficiency during large-scale evaluations. For the RAG retriever, we adopt the exact RAG configuration and textbook database from Patho-AgenticRAG \cite{patho-agenticrag}. For the tool-box, we implement 5 tools in total, including HoverNet \cite{graham2019hover} and CellViT++ \cite{horst2026cellvit++} for cell segmentation, CONCH \cite{conch} and Patho-CLIP \cite{patho-r1} for patch classification, and Patho-R1 \cite{patho-r1} for VQA. Crucially, to ensure strict separation between exploration and evaluation, the experience system is exclusively accumulated on the PathMMU val set during an initial exploration phase. 

The PathoSage framework operates in a hybrid deployment environment. The specialized tool models (HoverNet, CellViT++, CONCH, Patho-CLIP, and Patho-R1) and the RAG retrieval system are deployed locally on a computing node equipped with 8 $\times$ NVIDIA RTX 4090 GPUs. When utilizing proprietary models such as GPT-5.4 or Gemini-3-Pro as the host VLM, we directly access their respective cloud APIs. Conversely, for experiments evaluating the open-weights Qwen3-VL-32B-Instruct as the host VLM, we deploy the model locally using vLLM on an additional dedicated node equipped with 8 $\times$ NVIDIA RTX 4090 GPUs, utilizing tensor parallelism (TP=8) to ensure efficient inference.

\subsection{Prompts}
\begin{figure*}[t!]
  \centering
   \includegraphics[width=1.\textwidth]{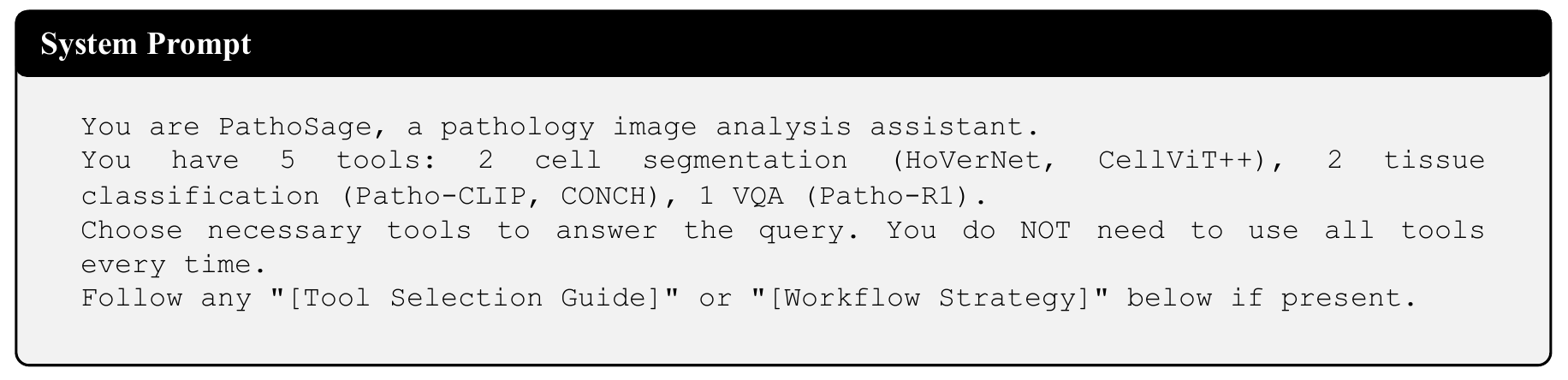}
   \caption{System prompt for PathoSage.}
   \label{fig_sys_prompt}
\end{figure*}

\begin{figure*}[t!]
  \centering
   \includegraphics[width=1.\textwidth]{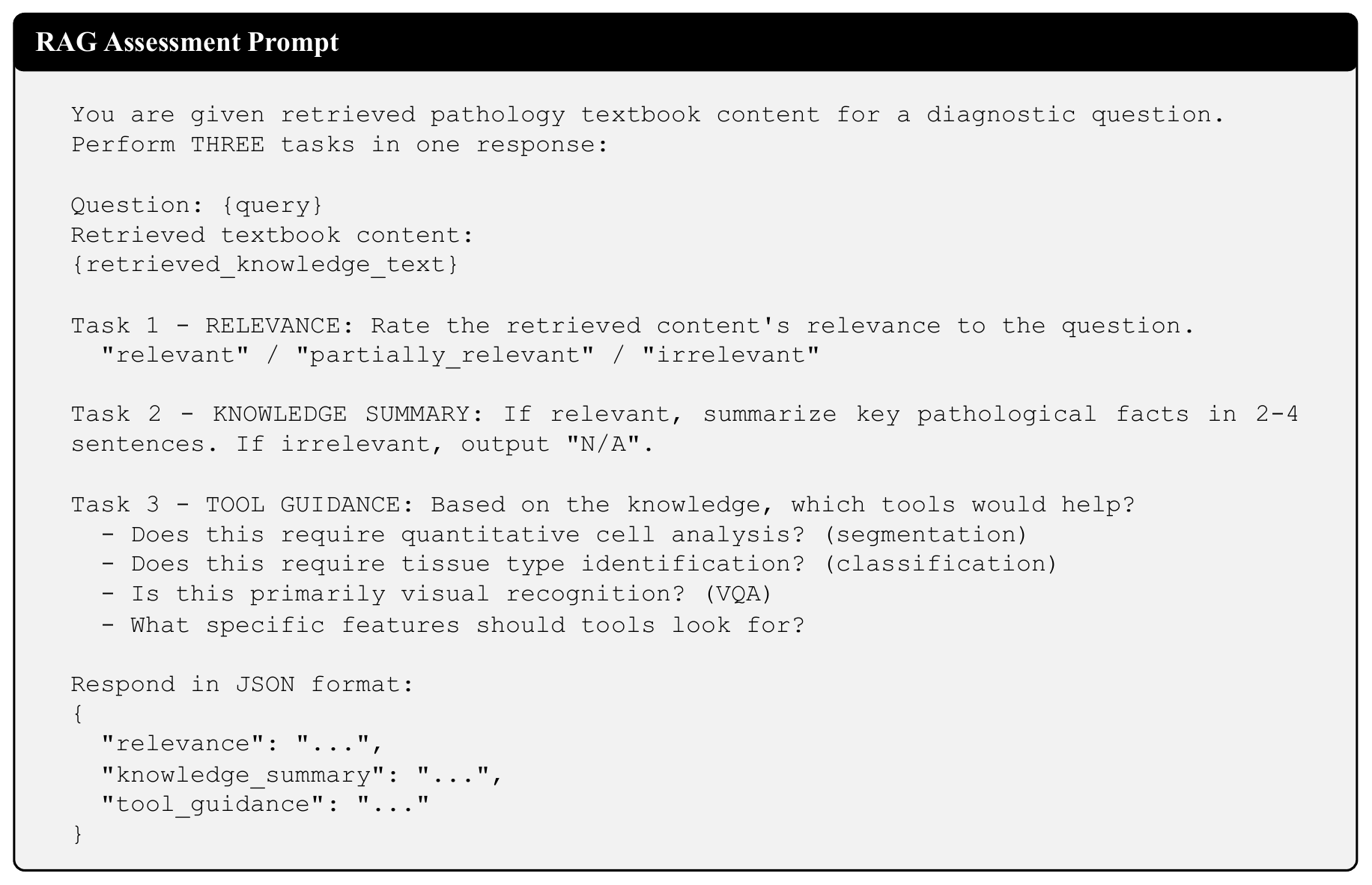}
   \caption{RAG assessment prompt for evaluating RAG retrieval results and generating recommendations for using tools.}
   \label{fig_rag_prompt}
\end{figure*}

\begin{figure*}[t!]
  \centering
   \includegraphics[width=1.\textwidth]{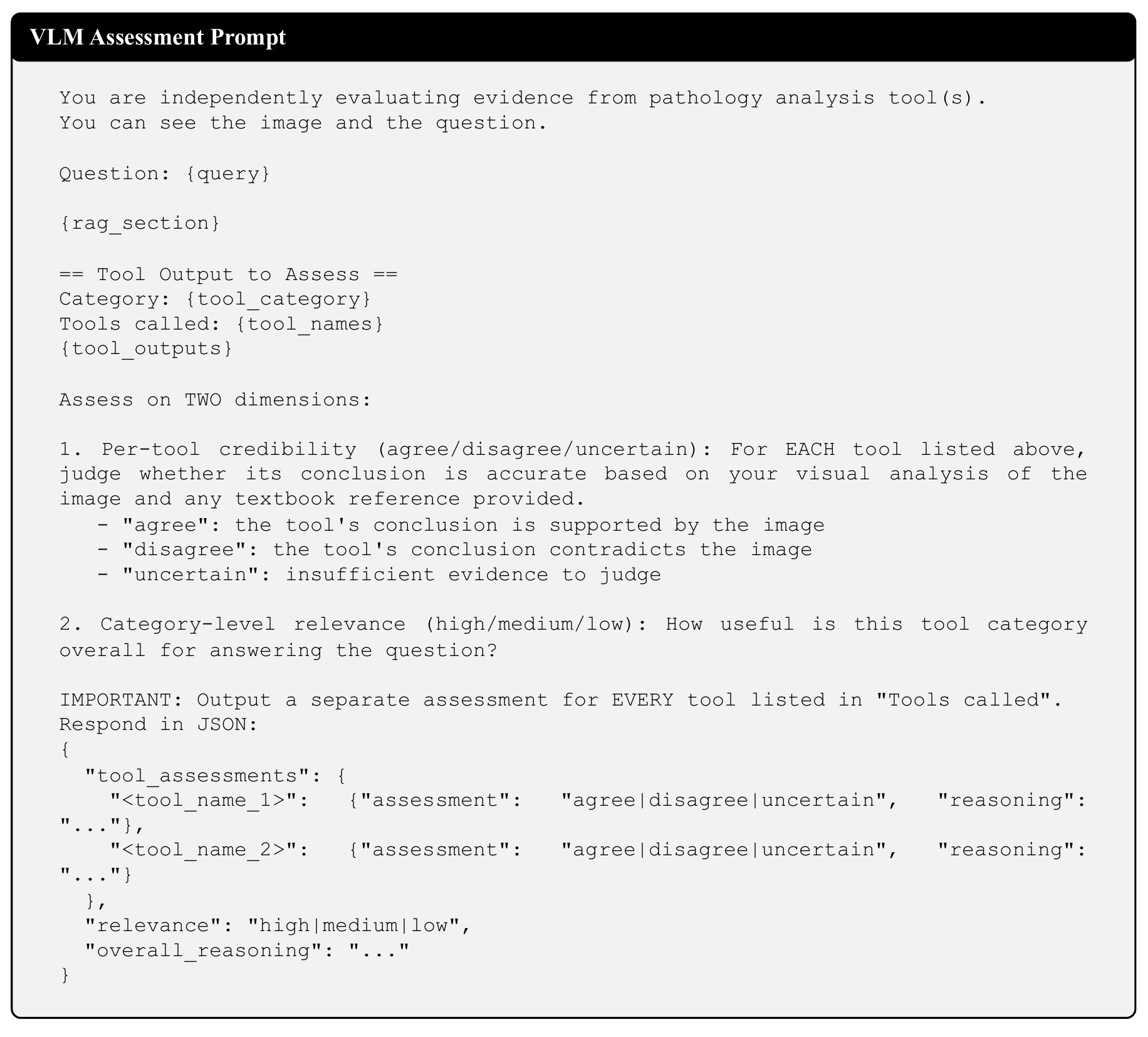}
   \caption{VLM assessment prompt for assigning independent scores to each tool's output.}
   \label{fig_vlm_assess_prompt}
\end{figure*}

\begin{figure*}[t!]
  \centering
   \includegraphics[width=1.\textwidth]{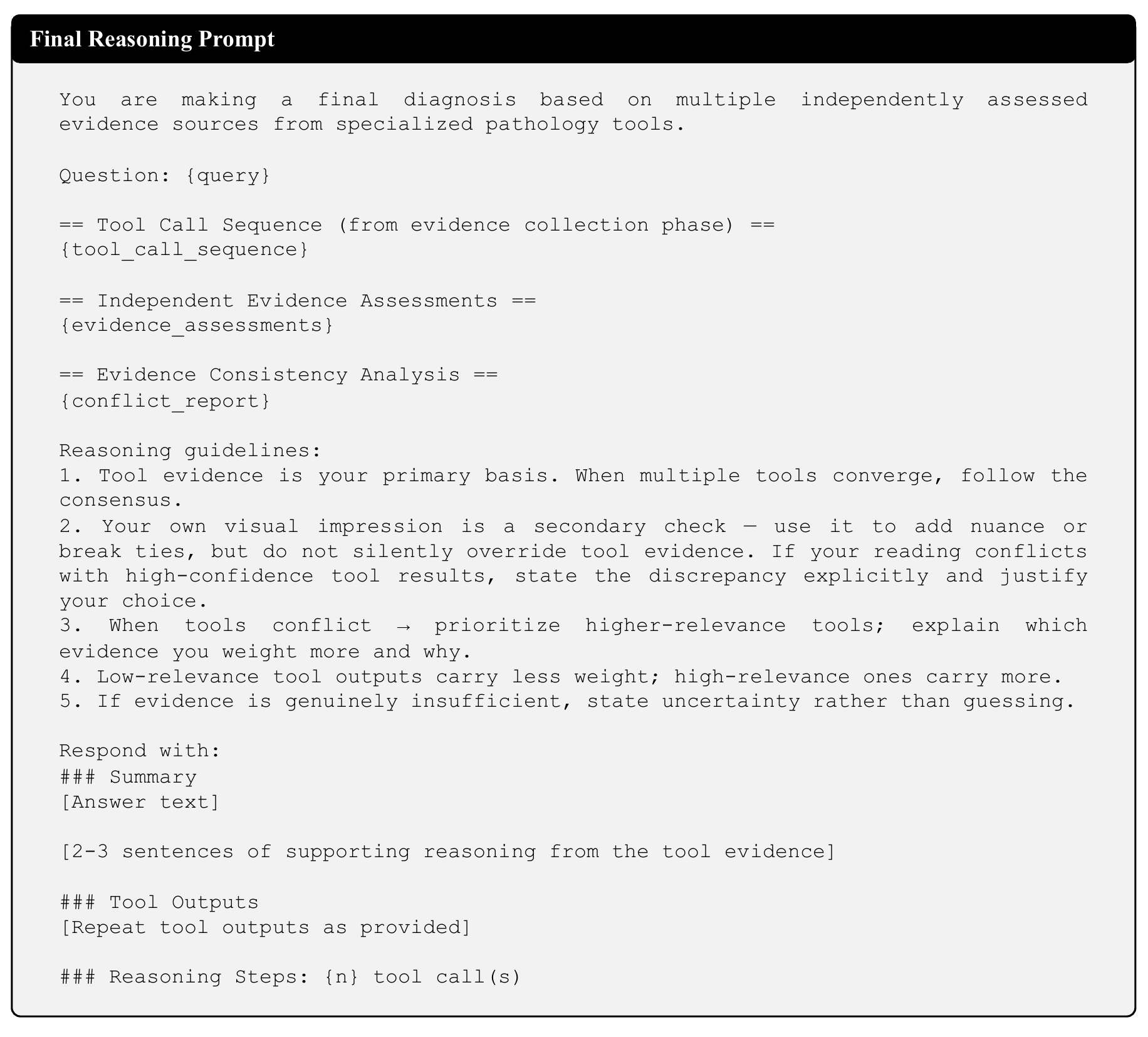}
   \caption{Final reasoning prompt for generating the output.}
   \label{fig_final_reasoning_prompt}
\end{figure*}

Figures \ref{fig_sys_prompt}–\ref{fig_final_reasoning_prompt} present the four prompts that govern the key decision points of PathoSage. Figure \ref{fig_sys_prompt} illustrates the system prompt, which functions as the system message during the ReAct-based evidence collection phase. It defines the agent’s role, enumerates the available tools, and specifies the tool-selection policy. Figure \ref{fig_rag_prompt} depicts the RAG assessment prompt, which is employed during the knowledge retrieval phase prior to tool invocation. This prompt instructs the model to evaluate the relevance of retrieved textbook passages, summarize key pathological knowledge, and translate this information into actionable guidance for tool selection in the subsequent ReAct phase. Figure \ref{fig_vlm_assess_prompt} shows the independent assessment prompt used in Step 1 of the SED phase, which elicits a credibility label (agree, disagree, or uncertain) for each executed tool and a relevance label (high, medium, or low) for each tool category. Figure \ref{fig_final_reasoning_prompt} presents the final reasoning prompt, applied in the last step of the SED phase, which integrates per-tool assessments and cross-tool conflict reports to generate the final diagnostic output under explicit evidence-weighting rules.

\end{document}